\documentclass[10pt,twocolumn,letterpaper]{article}

\usepackage{wacv}              

\usepackage{graphicx}
\usepackage{amsmath}
\usepackage{amssymb}
\usepackage{booktabs}
\usepackage{comment} 
\usepackage{xcolor}
\usepackage{colortbl}
\usepackage{multirow}
\usepackage{subcaption}

\usepackage[pagebackref,breaklinks,colorlinks]{hyperref}

\usepackage[capitalize]{cleveref}
\crefname{section}{Sec.}{Secs.}
\Crefname{section}{Section}{Sections}
\Crefname{table}{Table}{Tables}
\crefname{table}{Tab.}{Tabs.}

\def\wacvPaperID{1} 
\def\confName{WACV}
\def\confYear{2025}

\begin{document}

\title{SkipClick: Combining Quick Responses and Low-Level Features for Interactive Segmentation in Winter Sports Contexts}

\author{Robin Schön \and Julian Lorenz \and Daniel Kienzle \and Rainer Lienhart \\
University of Augsburg\\
Germany, 86159 Augsburg, Universitätsstr. 6a\\
{\tt\small \{robin.schoen, julian.lorenz, daniel.kienzle, rainer.lienhart\}@uni-a.de}
}

\maketitle

\begin{abstract}
In this paper, we present a novel architecture for interactive segmentation in winter sports contexts. 
The field of interactive segmentation deals with the prediction of high-quality segmentation masks by informing the network about the objects position with the help of user guidance. In our case the guidance consists of click prompts. 
For this task, we first present a baseline architecture which is specifically geared towards quickly responding after each click. 
Afterwards, we motivate and describe a number of architectural modifications which improve the performance when tasked with segmenting winter sports equipment on the WSESeg dataset. 
With regards to the average NoC@85 metric on the WSESeg classes, we outperform SAM and HQ-SAM by 2.336 and 7.946 clicks, respectively. When applied to the HQSeg-44k dataset, our system delivers state-of-the-art results with a NoC@90 of 6.00 and NoC@95 of 9.89. 
In addition to that, we test our model on a novel dataset containing masks for humans during skiing. \footnote{We will release code and the dataset  upon publication: \url{https://github.com/Schorob/skipclick}}
\end{abstract}

\section{Introduction} 
\begin{figure}
    \centering
    \includegraphics[width=0.95\linewidth]{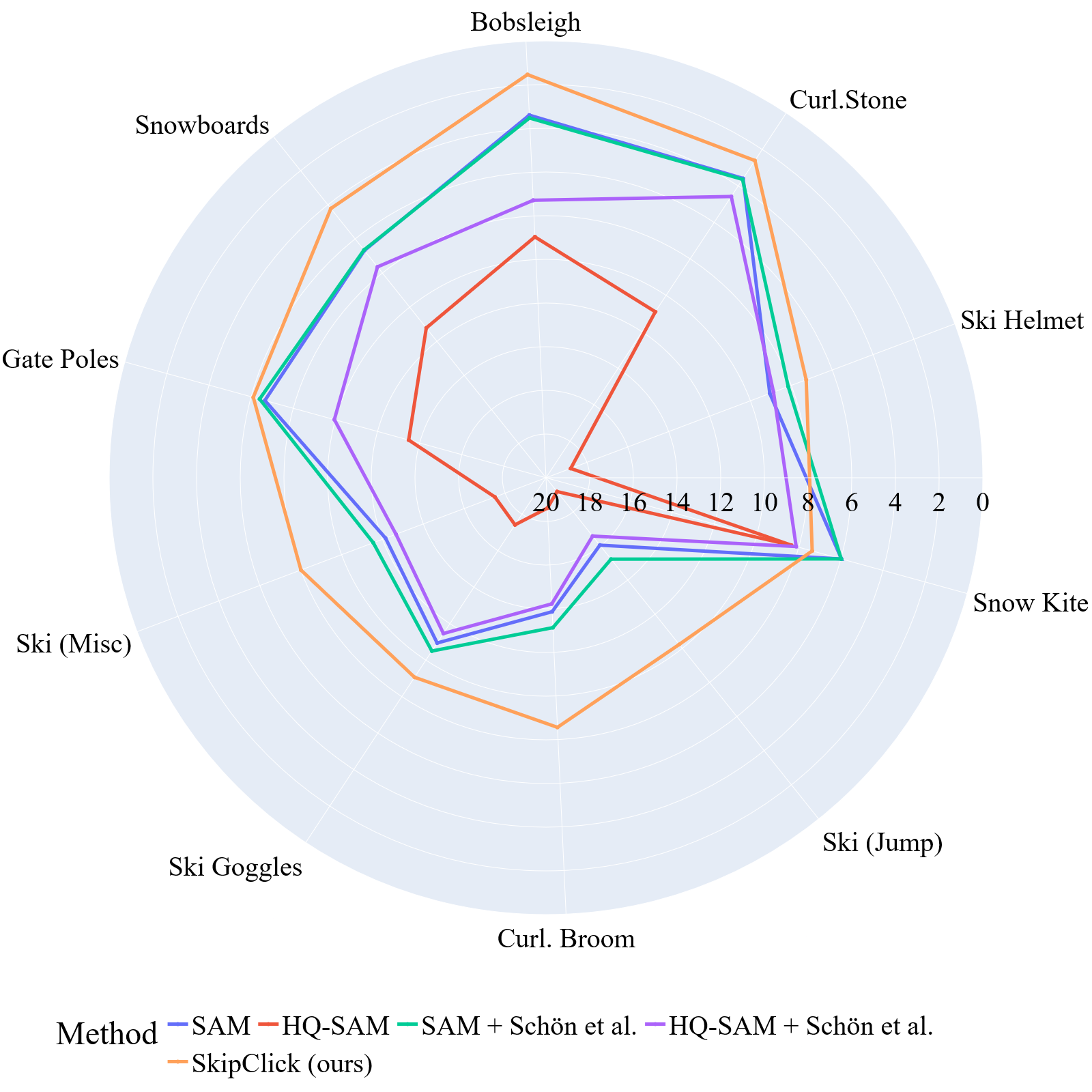}
    \caption{A comparison of the performance of SkipClick with other methods on the WSESeg \cite{schoen2024wseseg} dataset. The metric is NoC@85. }
    \label{fig:radar_plot_comparison}
\end{figure}
One of the most important applications of computer vision to sports, and therefore to winter sports as well, is the exact localization of athletes and objects involved in the activity. 
A large part of the attention falls upon the localization of the athlete. This mostly happens in the form of pose estimation \cite{ludwig2021self} and athlete segmentation \cite{Cioppa2019ARTHuS, Li2017MultipleHumanPI} . 
Recently, however, there has been an increased focus on the localization of the equipment used in the sport activities. 
In some cases this takes the form of segmenting the sports equipment \cite{sawahata2024instance, deepsportsradarv2} while other papers make use of the segmentation mask for various additional purposes \cite{ludwig2023all, ludwig2023detecting}. 
This development entails an increased need for annotated data in this direction. On top of this, segmentation masks take a lot of time to annotate.
Especially with regards to annotation tools that allow the user to annotate polygons or use brush / eraser tools, annotating fine details can be very challenging when annotation speed also constitutes an important factor. The COCO dataset \cite{lin2014microsoft} for example is subject to slight errors in the annotated segmentation masks. 
For this reason, interactive segmentation systems have been developed. The purpose of these systems lies in aiding the user with segmenting an image by trying to infer high quality image segmentation masks from used guidance. The main goal is to enable the user to produce such desired object segmentation masks in much less time than would be necessary to annotate manually from scratch. 
In most cases this amounts to clicks and scribbles to indicate the foreground / background on the images, or bounding boxes around the object that the user wants to segment. 
In recent years, neural network models have become the main component of such interactive segmentation systems. In these cases the network is given the image, the user guidance and in some cases a coarse mask of yet insufficient quality. The network is then tasked with predicting a segmentation mask for a desired object on the image, as indicated by the user guidance. 

In this paper we will only focus on iterative mask refinement with the help of foreground and background clicks: The user repeatedly inspects a mask and places a click on an erroneously segmented area. The segmentation network then uses the image, previous mask and the user clicks to predict an improved mask. This process is repeated until the user considers the object to be segmented adequately. 

Most earlier frameworks feed the clicks and previous mask to the model in conjunction with the image. In this situation which we call \emph{early fusion}, the early layers of the segmentation model have access to the prompts. Whilst generally allowing for improved mask quality, this entails a drawback with respect to the response time after each click: The entire model has to be run again in order to generate a new mask. 
One of the most notable developments is the Segment Anything Model \cite{kirillov2023segment}. In order to increase the interaction speed, SAM first computes a prompt-independent image feature tensor. This features tensor is then fused with the prompt information by two lightweight networks: a prompt encoder and a mask decoder. Whenever the user adds a new click, only the two lightweight networks have to be rerun, allowing for a quick response of the system. 
We refer to this as \emph{late fusion}. 
Since SAM is trained on the SA-1B dataset, which contains 1.1 billion masks, it constitutes a viable foundation model. 
The authors of \cite{schoen2024wseseg}, however, mention a crucial problem of SAM. Despite being trained on the largest dataset for interactive segmentation to date, it does not generalize well to the domain of winter sports equipment. The authors demonstrate this on their new WSESeg dataset. 

In this paper we will introduce a \textbf{new architecture for interactive segmentation in unusual domains} such as winter sports equipment, which we will call \emph{SkipClick}. 
The first and most important requirement of our model will be a \textbf{quick response time}. Secondly, we have to make sure to \textbf{avoid overfitting} since snow landscapes are a rather unusual domain in comparison to the regular consumer photos that constitute most available datasets. On top of this, we have to \textbf{explicitly enable the network to deal with fine structures} that sometimes occur when segmenting winter sports equipment. 
In addition to constructing a model that adheres to these requirements, we will be \textbf{able to demonstrate that our model is not over-engineered to the winter sports domain}. When compared to other real-time interactive segmentation models, our model shows competitive performance on datasets that contain only regular consumer photos.  
Since \cite{schoen2024wseseg} focuses on winter sports equipment instead of the athletes themselves, \textbf{we propose a new dataset with masks for skiers}. This new dataset is called SHSeg (\underline{S}kiing \underline{H}uman \underline{Seg}mentation) and contains 534 segmentation masks on 496 images. This new dataset will be used as an evaluation dataset. 

Our contributions can be summarized as follows: 
\begin{itemize}
    \item We present a real-time interactive segmentation model which is capable of performing segmentation on unusual domains such as winter sports. When designing the architecture, we pay explicit attention to enable the model to deal with fine-grained structures in the images. 
    \item We provide an ablation study of our model on the winter sports equipment segmentation dataset WSESeg. We are able to show that our model even surpasses the performance of SAM, despite being trained on the vastly smaller COCO+LVIS dataset. 
    \item We are also able to show that our model is not over-engineered to the winter sports domain. The model exhibits competitive or even superior performance on general consumer image datasets when compared to other late fusion based models.
    \item We present a novel segmentation dataset with masks for athletes during skiing. 
\end{itemize}

\section{Related Work}
\subsection{Usage of Segmentation Masks in Sports}
In \cite{Cioppa2019ARTHuS}, the authors propose the segmentation of participants of soccer and basketball matches. 
The authors of \cite{ghasemzadeh2021deepsportlab} present a system for segmenting players and localizing the ball conjointly, while \cite{gao2023sparse} focuses on the player localization aspect. 
The authors of \cite{huang2022parallel} also segment people in sports contexts, but include the used equipment in their masks. 
The method proposed in \cite{tarashima2021sports} aims at directly segmenting the basketball field itself. Similarly, the authors of \cite{liu2019instance} segment the fields in outdoor sports.
\cite{sawahata2024instance} track the sword in fencing matches by localizing the tip and segmenting the sword. 
The authors of \cite{deepsportsradarv2} estimate the surface of basketballs by estimating their location and radius. The authors of \cite{ludwig2023all, ludwig2023detecting} use the segmentation masks of skis to enable the detection of additional keypoints. 
\cite{schoen2024wseseg} is the most relevant publication to our work. Therein, the authors publish a novel dataset for the segmentation of winter sports equipment and techniques for the online adaptation of interactive segmentation methods.

\subsection{Interactive Segmentation}
Interactive segmentation deals with problem of producing high quality image segmentation masks by integrating information from user interactions. 
While there are certain methods which make use of hand-crafted image features and pixel values \cite{starconvexity2010, interactivegraphcuts2001,grabcut2004,hahn2003iwt}, most modern methods employ neural networks to perform this task. 
The most prevalent form of interaction are clicks that indicate single foreground and background pixels. 
The authors of \cite{ritm2022} discuss various different networks and the effectiveness of using the previous step's mask as input. 
The work in \cite{Liu_2023_ICCV} introduces transformers as an effective choice of backbone for interactive segmentation. 
The methods presented in \cite{jang2019interactive,sofiiuk2020f} use backpropagation to enforce consistency between the user interaction and the mask outputs. 
Most of the aforementioned methods provide masks of great quality at the cost of a slow response after each click. 
The authors of \cite{kirillov2023segment} introduce the concept of late fusion, provide the large SA-1B dataset and publish the weights to be used as a foundation model. 
\cite{ke2024segment} introduces the HQSeg-44k dataset and fine tunes SAM to increase the level of detail in the predicted masks. 
\cite{huang2023interformer} provides another architecture using late fusion to achieve an efficient response time. 
The baseline we arrive at in \cref{sec:method_construction} resembles \cite{liu2024rethinking} to a certain degree. 
There also has been an effort to adapt interactive segmentation systems during usage, as can be seen in \cite{lenczner2020interactive,schoen2024wseseg,schon2024adapting,lenczner2022dial}.

\section{Method}
\begin{figure*}
\centering
\begin{subfigure}{0.31\textwidth}
    \includegraphics[width=\linewidth]{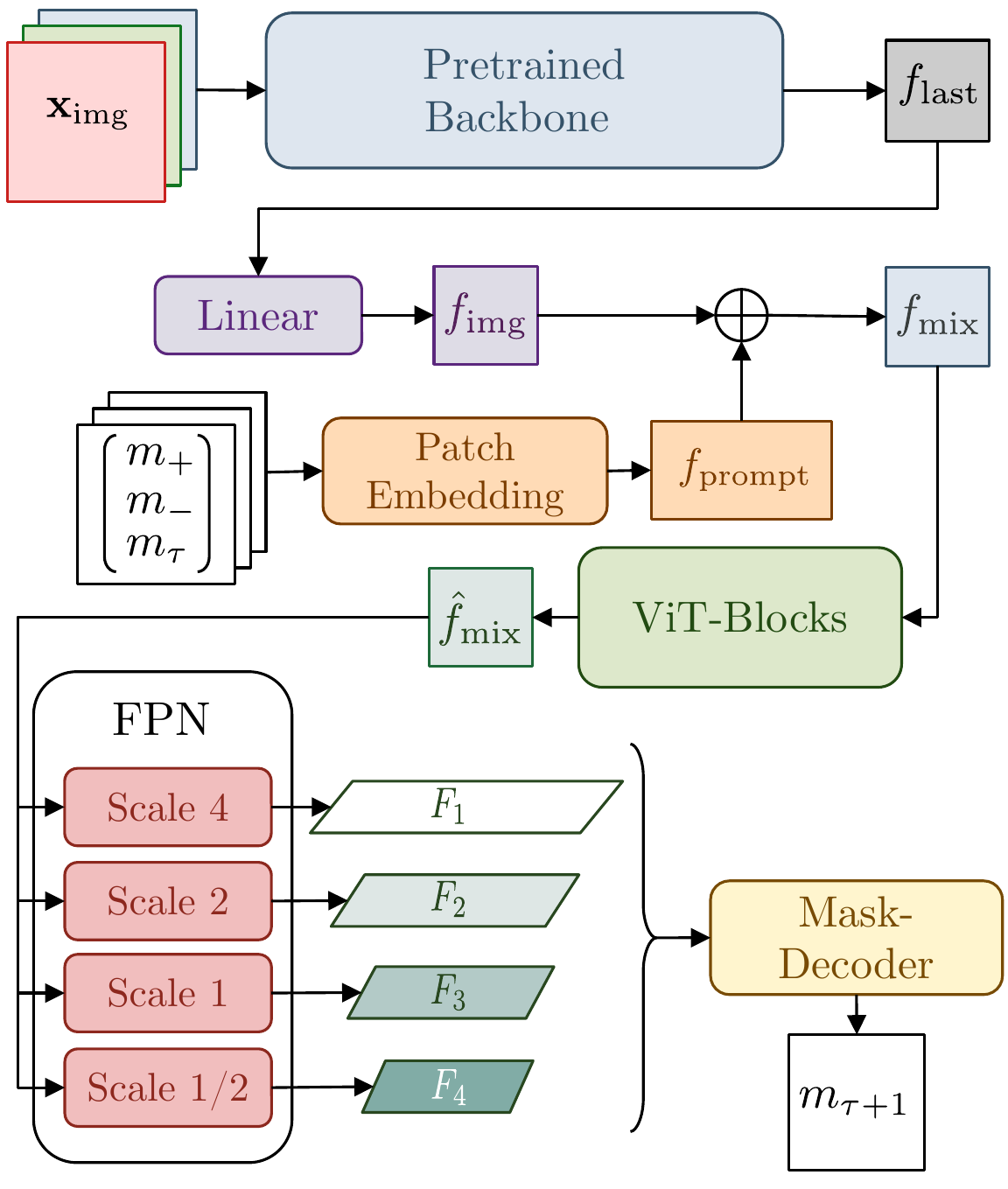}
    \caption{The baseline architecture. }
    \label{fig:baseline}
\end{subfigure}
\begin{subfigure}{0.67\textwidth}
    \includegraphics[width=\linewidth]{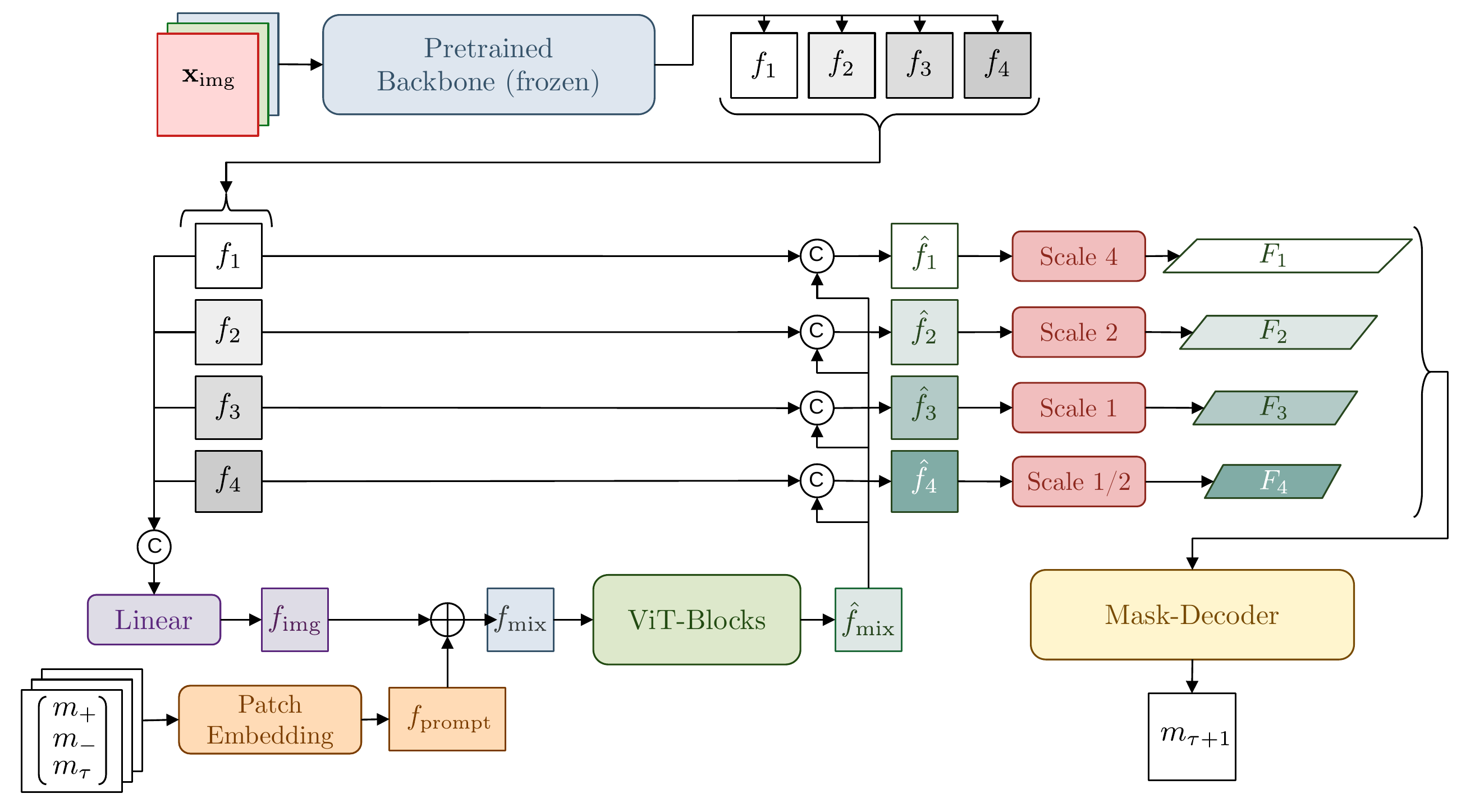}
    \caption{Our full SkipClick architecture. }
    \label{fig:architecture}
\end{subfigure}
\caption{The baseline (\emph{left}) and the SkipClick (\emph{right}) architecture. Note that the bulk of the computation happens in the backbone, which only has be executed once per image. Freezing the backbone during training enables the backbone to retain its generality from unsupervised pretraining. The use of multi-level features and skip connections allows the model to deal with fine structures encountered when segmenting winter sports equipment.}
\end{figure*}

We will first describe the problem of interactive segmentation by presenting how the task is usually structured. Afterwards we will describe our architecture in two stages. 
In the first stage, we will present a baseline based on certain requirements: A quick response time after the clicks, a shallow decoder and a backbone with a strong capacity for generalization. This baseline can be found in \cref{fig:baseline}. 
In the second stage, we present certain architectural modifications and show that they improve the performance of the model. Due to a lack of large scale winter sports related segmentation masks, we want the model be able to generalize well. Since we aim to segment winter sports equipment, which often has a surface with delicate structures, we pay special attention to enable our model to segment fine details by using low-level image feature tensors. 
The resulting architecture can be found in \cref{fig:architecture}. \Cref{tab:wseseg_ablation} displays the measured improvements.

\subsection{Problem Statement} 
\label{subsec:problem}
In interactive segmentation, we have an image $\mathbf{x}_\text{img} \in \mathbb{R}^{H \times W \times 3}$ which contains an object of interest. We want to create a high-quality segmentation mask $\mathbf{m} \in \{0, 1\}^{H \times W}$ which has the value 1 for all pixels belonging to our object of interest and the value 0 for all other pixels. We thus effectively want to distinguish foreground from background.
In order to achieve this, the user repeatedly interacts with a segmentation network $\mathcal{F}_\text{ISeg}$. Each of these interactions has the following structure: 
\begin{enumerate}
    \item During the $\tau$-th interaction the user inspects the current mask $\mathbf{m}_\tau$ for erroneous areas. In the beginning of the first iteration the initial mask $\mathbf{m}_0$ only consists of 0s (all background). 
    \item In order to give the network additional information, the user places a click $\mathbf{p}_{\tau} = (i_{\tau}, j_{\tau}, l_\tau)$ on the image. Here, $(i_{\tau}, j_{\tau}) \in \{1, ..., H\} \times \{1, ..., W\}$ is an image coordinate. Whether the clicked pixel belongs to the foreground (+) or background (-) is indicated by the label $l_\tau \in \{+,-\}$. The user sets the label by choosing to place the click either with the left or right mouse button. 
    \item The image $\mathbf{x}_\text{img}$, the current mask $\mathbf{m}_\tau$, and all accumulated clicks $\mathbf{p}_{0:\tau}$ so far are given to the network in order to produce an improved mask $\mathbf{m}_{\tau+1} = \mathcal{F}_\text{ISeg}(\mathbf{x}_\text{img}, \mathbf{m}_\tau, \mathbf{p}_{0:\tau})$.
\end{enumerate} 
Since the user gives an increasing amount of information to the network, we assume this method to offer the possibility to eventually arrive at a mask of very high quality. A good interactive segmentation system is thus characterized by minimizing the number of clicks necessary to create such a high quality mask. 

\subsection{Architecture} 
\label{sec:method_construction}
We will first describe a baseline architecture, which is held as simple as possible while still fulfilling a certain range of requirements. By doing so, we arrive at an architecture that bears some similarity to \cite{liu2024rethinking}. Their method also uses attention to integrate prompts, a ViT encoder and the mask predictor follows \cite{Liu_2023_ICCV}. However, in contrast to \cite{liu2024rethinking}, we use full transformer encoder blocks to fuse the features and refrain from using cross attention.
After introducing the baseline, we look at some modifications and justify each of them. This will eventually lead to our complete SkipClick architecture.
\paragraph{The baseline.} 
We want to offer a simple baseline architecture, which generalizes well and equips the network with the capability to deliver real-time responses after each click. 
As a consequence we aim for the module which combines the interaction information and the image features to remain relatively shallow. Apart from the mask predictor, where we follow common practice, our baseline will be an almost direct consequence of the aforementioned requirements. 

The first requirement is a large potential for generalization, since we want to be able to segment a vast variety of shapes in various different contexts. Here we will follow common practice and use a vision transformer architecture \cite{kirillov2023segment, Liu_2023_ICCV}. Interactive segmentation is generally class agnostic in nature. Thus, we do not want our backbone to be constrained by the bias induced when pre-training on a labeled dataset. We will opt for a ViT \cite{DBLP:conf/iclr/DosovitskiyB0WZ21} that has been pre-trained with the DINOv2 framework \cite{oquab2023dinov2}, since this pretext task does not make use of labels. In our baseline, the input image is transformed into an image feature tensor 
\begin{equation}
\label{eq:fimg_base}
    f_\text{img} = \text{Linear}(\text{ViTBackbone}(\mathbf{x}_\text{img})) \in \mathbb{R}^{\frac{H}{14} \times \frac{W}{14} \times d_\text{model}}, 
\end{equation}
where $H, W$ and $d_\text{model}$ are the image height, image width and the feature dimension in the transformer, respectively.
The second requirement consists in the desire to get a real-time response from the neural network. Until recently, interactive segmentation methods reran the entire network after each click \cite{ritm2022, Liu_2023_ICCV, chen2022focalclick}. Therein, the input image was fused with the previous mask and the foreground / background clicks very early in the network. This had the side effect of having to repeat the bulk of the computation after every interaction. While these types of network usually perform really well with respect to the number of clicks, the slow response time is a hindrance to the user during actual usage. 
Recent publications, such as SAM \cite{kirillov2023segment} and InterFormer \cite{huang2023interformer}, have tackled this problem by splitting the network into a heavyweight image encoder and a lightweight mask predictor. The heavyweight image encoder constitutes the bulk of the networks computation and only needs to be run once, since the image itself never changes during the segmentation interactions. After each interaction, the lightweight mask predictor is given the image features, the previous mask and the encoded clicks in order to predict an improved version of the mask. Since only the lightweight mask predictor has to be executed after each interaction, the response time the user experiences diminishes drastically. We will adopt this strategy of only encoding the image once, and executing a lightweight network after each interaction. 

We combine the image features $f_\text{img}$ with the user guidance as follows:
We encode the positive and negative clicks in $\mathbf{p}_{0:\tau}$ by drawing small disks with a radius of 5 pixels on two different binary images. This results in two click maps $\mathbf{m}_+, \mathbf{m}_- \in \{0,1\}^{H \times W}$. These click maps are then concatenated with the mask from the previous round $\mathbf{m}_\tau$ and fed to a patch embedding layer to obtain 
\begin{equation}
    f_\text{prompt} = \text{PatchEmbedding}(\text{Concat}(\mathbf{m}_+, \mathbf{m}_-, \mathbf{m}_\tau])),  
\end{equation}
where the patch size is $14 \times 14$ as in DINOv2. 
In order to combine the image features and the prompt features, we add the two tensors in an element-wise fashion to obtain 
\begin{equation}
    f_\text{mix} = f_\text{img} + f_\text{prompt}. 
\end{equation}

Since we aim for a lightweight architecture, we will need it to be relatively shallow. Because we want to guarantee for a sufficiently large field of view, whilst stacking the models in a shallow fashion, we use transformer blocks to incorporate the image features and the interaction information. This results in 
\begin{equation}
    \hat{f}_\text{mix} = \text{ViTBlocks}(f_\text{mix}). 
\end{equation}
When it comes to the mask head, we will use the same head as SimpleClick: First the feature tensor is fed to an feature pyramid network (FPN) as described in \cite{li2022exploring}. This feature pyramid is then given to a SegFormer decoder \cite{xie2021segformer} to obtain 
\begin{equation}
    \mathbf{m}_{\tau+1} = \text{MaskDecoder}(\text{FPN}(\hat{f}_\text{mix})).
\end{equation}

It should be noted that our baseline is similar to the method proposed in \cite{liu2024rethinking}, with two main differences. First, we use transformer blocks and not just the attention mechanism to incorporate interactions and image features. Second, we do not use any form of cross-attention or text integration.

\paragraph{SkipClick.}

\begin{table*}[t]
    \centering
    \resizebox{\linewidth}{!}{
    
    \begin{tabular}{|c|c|c|c|c|c|c|c|c|c|c|c|}
        \hline
         \multicolumn{2}{|l|}{\multirow{2}{*}{\textbf{Configuration}}}  & \multicolumn{2}{c|}{Bobsleigh} & \multicolumn{2}{c|}{Curl. Stone}  & \multicolumn{2}{c|}{Ski Helmet} & \multicolumn{2}{c|}{Snow Kite} & \multicolumn{2}{c|}{Ski (Jump)}   \\
         
         \multicolumn{2}{|l|}{} & NoC@85 & NoC@90 & NoC@85 & NoC@90 & NoC@85 & NoC@90 & NoC@85 & NoC@90 & NoC@85 & NoC@90 \\
         \hline
         \multicolumn{2}{|l|}{Baseline} & 4.48 & 6.15 & 6.53 & 8.29 & 7.95 & 10.61 & 9.1 & 10.82 & 15.05 & 18.26 \\
         \multicolumn{2}{|l|}{+ Frozen Backbone} & 4.01 & 5.64 & 6.11 & 7.98 & 7.55 & 10.54 & 9.32 & 10.77 & 15.02 & 17.96 \\ 
         \multicolumn{2}{|l|}{+ Intermediate Features} & 1.6 & 2.19 & 3.21 & 4.99 & 8.22 & 10.82 & 7.5 & 9.56 & 11.5 & 16.43 \\
         \multicolumn{2}{|l|}{+ Skip Connections} & 1.52 & 2.08 & 2.61 & 4.09 & 7.27 & 9.31 & 7.36 & 8.93 & 10.23 & 15.51 \\
         \hline
         \multicolumn{2}{|c|}{Curl. Broom} & \multicolumn{2}{c|}{Ski Goggles} & \multicolumn{2}{c|}{Ski (Misc)} & \multicolumn{2}{c|}{Slalom Gate Poles} & \multicolumn{2}{c|}{Snowboards} &  \multicolumn{2}{c|}{\cellcolor[HTML]{DDDDDD}\textbf{Average}} \\
         NoC@85 & NoC@90 & NoC@85 & NoC@90 & NoC@85 & NoC@90 & NoC@85 & NoC@90 & NoC@85 & NoC@90 & NoC@85 & NoC@90 \\
         \hline
         13.55 & 16.42 & 9.14 & 12.38 & 12.17 & 15.03 & 9.4 & 13.12 & 7.26 & 9.23 & 9.463 & 12.031 \\ 
         14.02 & 16.62 & 8.9 & 12.33 & 12.36 & 15.11 & 9.9 & 13.65 & 6.97 & 8.91 & 9.416 & 11.951 \\ 
         10.63 & 15.12 & 8.97 & 12.38 & 9.23 & 13.02 & 7.24 & 12.19 & 4.75 & 6.74 & 7.285 & 10.344 \\ 
         8.56 & 13.16 & 9.06 & 11.5 & 8.01 & 11.49 & 6.1 & 9.62 & 4.22 & 5.94 & \textbf{6.494} & \textbf{9.163} \\
         \hline
    \end{tabular}
    }
    \caption{Results on the WSESeg dataset validating our architectural choices. The rows describe our architecture in a cumulative manner, meaning each row also contains the configurational aspects of the rows above. The final configuration is our SkipClick architecture. The lower the NoC metric, the better the performance. }
    \label{tab:wseseg_ablation}
    
\end{table*}

We will use our central ablation study to motivate the decisions in constructing our architecture in a step-by-step fashion. While doing so, we will show the improvements incurred by each design decision. Therefore, we recommend any reader unacquainted with interactive segmentation and the WSESeg dataset to first read \Cref{sec:setting}. The full architecture can be seen in \cref{fig:architecture} and the improvements incurred by the architectural elements in \cref{tab:wseseg_ablation}. 

The most important aspect of dealing with winter sports in machine learning settings is generalization. Especially winter sports equipment occurs rather scarcely in datasets made up of general consumer images. 
Although snowboards and skis happen to be part of the COCO dataset, their masks are often only coarse polygons of mediocre quality. 
Since the segmentation training data is implicitly limited to certain object types, we incur a risk of overfitting. Especially the backbone might loose some of its generality.
In order to alleviate this problem take a step that may seem rather counterintuitive. We freeze the backbone which is used to encode the image. 
Our results corroborate this decision: When comparing the first two lines in \cref{tab:wseseg_ablation}, we see an improvement of the average NoC@85 metric (defined in \Cref{sec:setting}) from 9.463 to 9.416.

Recent techniques \cite{Liu_2023_ICCV, kirillov2023segment, liu2024rethinking} rely on the last feature tensor computed by the ViT encoder to still contain enough fine-grained information about the image to perform pixel-level tasks such as segmentation. Especially in the winter sports domain, where some objects may have delicate details in their shapes, such fine-grained information is essential when producing novel segmentation masks. 
We assume the information from the features generated by the last layer to lack low-level details. We therefore use various intermediate feature tensors as well. In particular we use the feature tensors after the 3rd, 6th, 9th and 12th encoder block of the ViT-B, which we call $f_1, f_2, f_3$ and $f_4$, respectively. 
\begin{equation}
    f_1, f_2, f_3, f_4 = \text{ViTBackbone}(\mathbf{x}_\text{img})
\end{equation}
We integrate the information from various layers in the backbone by concatenating their output tensor and feeding them to a linear layer. This results in an alteration of \cref{eq:fimg_base} to compute 
\begin{equation}
    f_\text{img} = \text{Linear}(\text{Concat}(f_1, f_2, f_3, f_4)). 
\end{equation}
This change incurs a considerable improvement, lowering NoC@85 from 9.416 to 7.285, whilst only extending the baseline by a single linear layer. 

So far our network only has access to the fine-grained information \emph{before} the integration of the prompts happens. We do however want our network to be able to access the intermediate features tensors \emph{after} the prompts are integrated, since the prompts are necessary for the network to know what object we want to segment at all. Inspired by U-Net \cite{ronneberger2015u}, we choose to integrate skip connections (see \cref{fig:architecture}). This helps us in creating the intermediate tensors 
\begin{equation}
    \hat{f}_i = \text{Concat}(\hat{f}_\text{mix}, f_i) \qquad \text{for } i = 1,2,3,4. 
\end{equation}
Right now, we have $\hat{f}_i \in \mathbb{R}^{\frac{H}{14} \times \frac{W}{14} \times 2 \cdot d_\text{model}}$. In order to obtain a feature pyramid as in \cite{li2022exploring} the tensors are rescaled to $4\times, 2\times, 1\times$ and $\frac{1}{2} \times$ their size as show in \cref{fig:architecture}. The rescaling modules do not simply upsample/downsample, but instead use (transposed) convolutions and non-linearities as in \cite{Liu_2023_ICCV}. The resulting feature maps $F_1, F_2, F_3$ and $F_4$ are processed by a SegFormer decoder to obtain 
\begin{equation}
    \mathbf{m}_{\tau+1} = \text{MaskDecoder}(F_1, F_2, F_3, F_4). 
\end{equation}
Bridging low-level features to the feature pyramid network improves our model again, reducing the NoC@85 from 7.285 to 6.494. 
When looking at all the architectural changes in \cref{tab:wseseg_ablation} in total, we observe a drop of 2.969 in the NoC@85 metric and a drop of 2.868 in the NoC@90 metric.

\section{Experiments} 

\begin{table*}[t]
    \centering
    \resizebox{\linewidth}{!}{
    
    \begin{tabular}{|c|c|c|c|c|c|c|c|c|c|c|c|}
        \hline
         \multicolumn{2}{|l|}{\multirow{2}{*}{\textbf{Configuration}}}  & \multicolumn{2}{c|}{Bobsleigh} & \multicolumn{2}{c|}{Curl. Stone}  & \multicolumn{2}{c|}{Ski Helmet} & \multicolumn{2}{c|}{Snow Kite} & \multicolumn{2}{c|}{Ski (Jump)}   \\
         
         \multicolumn{2}{|l|}{} & NoC@85 & NoC@90 & NoC@85 & NoC@90 & NoC@85 & NoC@90 & NoC@85 & NoC@90 & NoC@85 & NoC@90 \\
         \hline
         \multicolumn{2}{|l|}{SAM \cite{kirillov2023segment}} & 3.38 & 5.33 & 3.59 & 6.26 & 9.05 & 12.70 & 5.97 & 8.23 & 16.05 & 18.43 \\ 
         \multicolumn{2}{|l|}{HQ-SAM \cite{ke2024segment}} & 8.96 & 11.83 & 10.90 & 13.44 & 18.79 & 19.58 & 8.34 & 11.04 & 19.19 & 19.88 \\ 
         \multicolumn{2}{|l|}{SAM + Schön et al. \cite{schoen2024wseseg}} & 3.51 & 5.45 & 3.64 & 6.39 & 8.15 & 12.41 & 5.98 & 8.21 & 15.23 & 18.42 \\ 
         \multicolumn{2}{|l|}{HQ-SAM + Schön et al. \cite{schoen2024wseseg} \textdagger} & 7.28 & 8.24 & 4.58 & 7.71 & 8.87 & 12.85 & 8.12 & 10.42 & 16.58 & 18.98 \\ 
         \hline 
         \multicolumn{2}{|l|}{SkipClick (Ours)} & 1.52 & 2.08 & 2.61 & 4.09 & 7.27 & 9.31 & 7.36 & 8.93 & 10.23 & 15.51 \\
         \hline
         \multicolumn{2}{|c|}{Curl. Broom} & \multicolumn{2}{c|}{Ski Goggles} & \multicolumn{2}{c|}{Ski (Misc)} & \multicolumn{2}{c|}{Slalom Gate Poles} & \multicolumn{2}{c|}{Snowboards} &  \multicolumn{2}{c|}{\cellcolor[HTML]{DDDDDD}\textbf{Average}} \\
         NoC@85 & NoC@90 & NoC@85 & NoC@90 & NoC@85 & NoC@90 & NoC@85 & NoC@90 & NoC@85 & NoC@90 & NoC@85 & NoC@90 \\
         \hline
         13.86 & 17.73 & 10.94 & 14.57 & 12.15 & 15.41 & 6.65 & 10.46 & 6.68 & 9.49 & 8.83 & 11.86 \\
         18.61 & 19.68 & 17.42 & 18.24 & 17.50 & 18.74 & 13.48 & 17.12 & 11.22 & 13.59 & 14.44 & 16.31 \\ 
         13.13 & 17.85 & 10.50 & 14.41 & 11.55 & 15.49 & 6.40 & 10.46 & 6.65 & 9.45 & 8.48 & 11.85 \\ 
         14.22 & 17.56 & 11.46 & 14.82 & 12.66 & 16.67 & 9.95 & 14.18 & 7.64 & 10.48 & 10.14 & 13.19 \\
         \hline
         8.56 & 13.16 & 9.06 & 11.5 & 8.01 & 11.49 & 6.1 & 9.62 & 4.22 & 5.94 & \textbf{6.494} & \textbf{9.163} \\
         \hline
    \end{tabular}
    }
    \caption{A comparison of SkipClick with previous methods on the WSESeg dataset. Most of the numerical results are taken from \cite{schoen2024wseseg}. \textdagger: When combining HQ-SAM with the method from \cite{schoen2024wseseg}, one of their ablations performs vastly better than their full method. For this reason we used the NoC metrics from their better performing ablation in this table row.}
    \label{tab:wseseg_comparison}
    
\end{table*}

\begin{table}[t]
    \centering

    \resizebox{0.80\linewidth}{!}{
    
    \begin{tabular}{|l|c|c|}
    \hline
    \textbf{Method} & NoC@85 & NoC@90 \\
    \hline 
    SAM \cite{kirillov2023segment} & 3.95 & 7.46 \\ 
    HQ-SAM \cite{ke2024segment} & 10.60 & 14.29 \\ 
    SAM + Schön et al. \cite{schoen2024wseseg} & 3.88 & 7.47 \\ 
    HQ-SAM + Schön et al. \cite{schoen2024wseseg} \textdagger & 4.70 & 8.45 \\ 
    \hline 
    SkipClick (Ours) & \textbf{1.44} & \textbf{2.52} \\
    \hline
    \end{tabular}
    }
    \caption{A comparison of SkipClick with previous methods on our newly proposed SHSeg dataset. \textdagger: Instead of taking the full method, we take the ablation that delivered the best performance on WSESeg in \cite{schoen2024wseseg}, as we did in \cref{tab:wseseg_comparison}. }
    \label{tab:shseg_comparison}
\end{table}

\begin{figure}
    \centering
    {\def\plotwidth{0.40\linewidth}
    \setlength{\tabcolsep}{1.5pt}
    \begin{tabular}{cc} 
         \textbf{Prediction} & \textbf{Ground Truth} \\
         \includegraphics[width=\plotwidth]{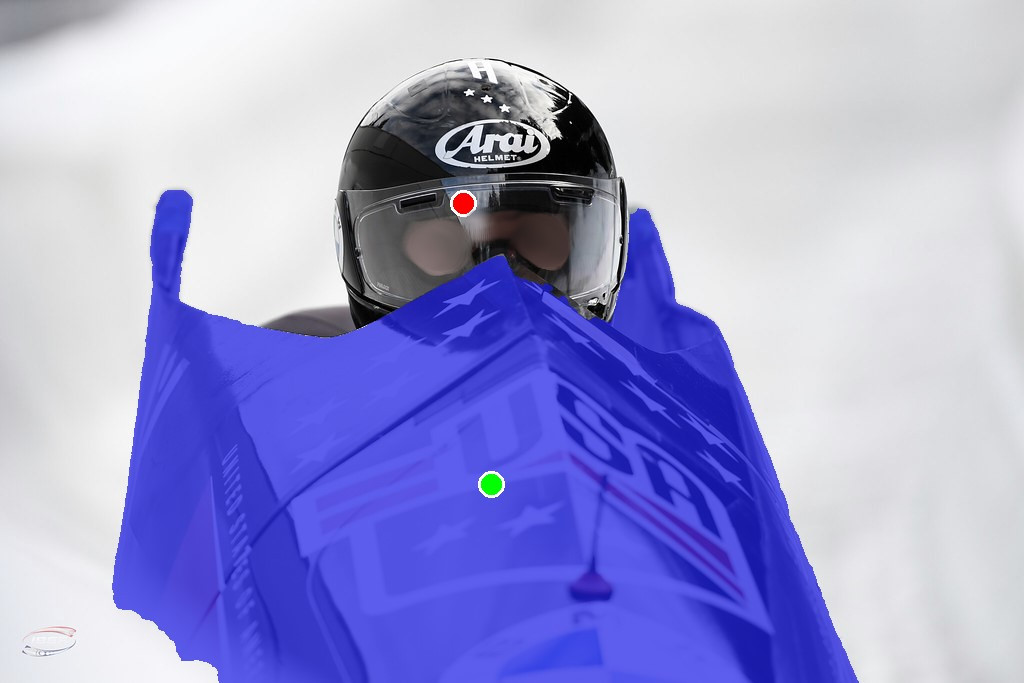} & 
         \includegraphics[width=\plotwidth]{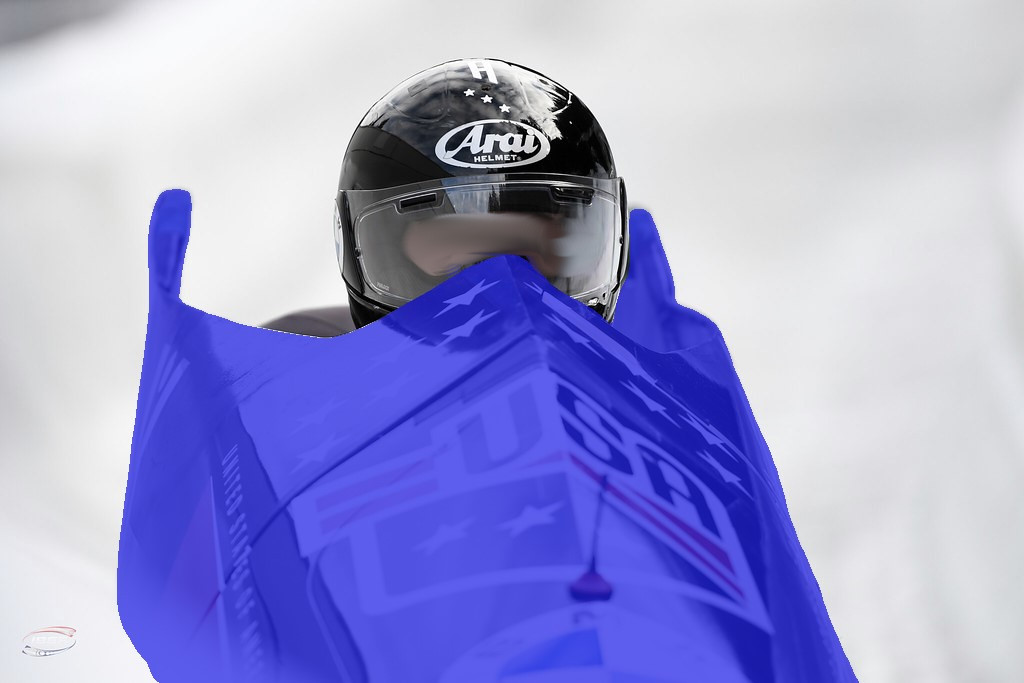} \\
         \includegraphics[width=\plotwidth]{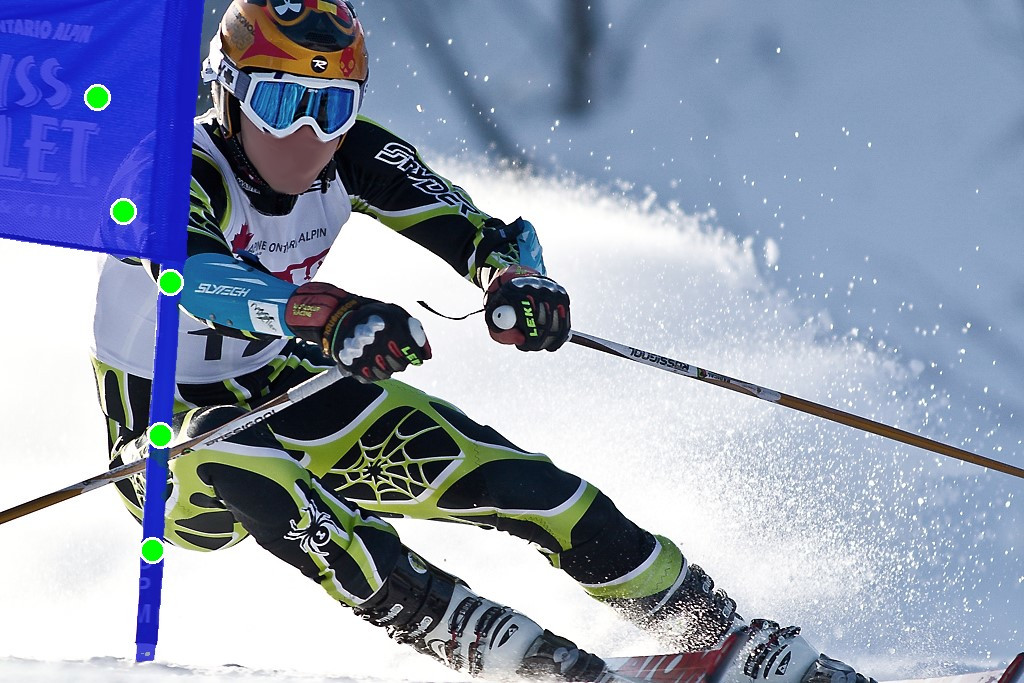} & 
         \includegraphics[width=\plotwidth]{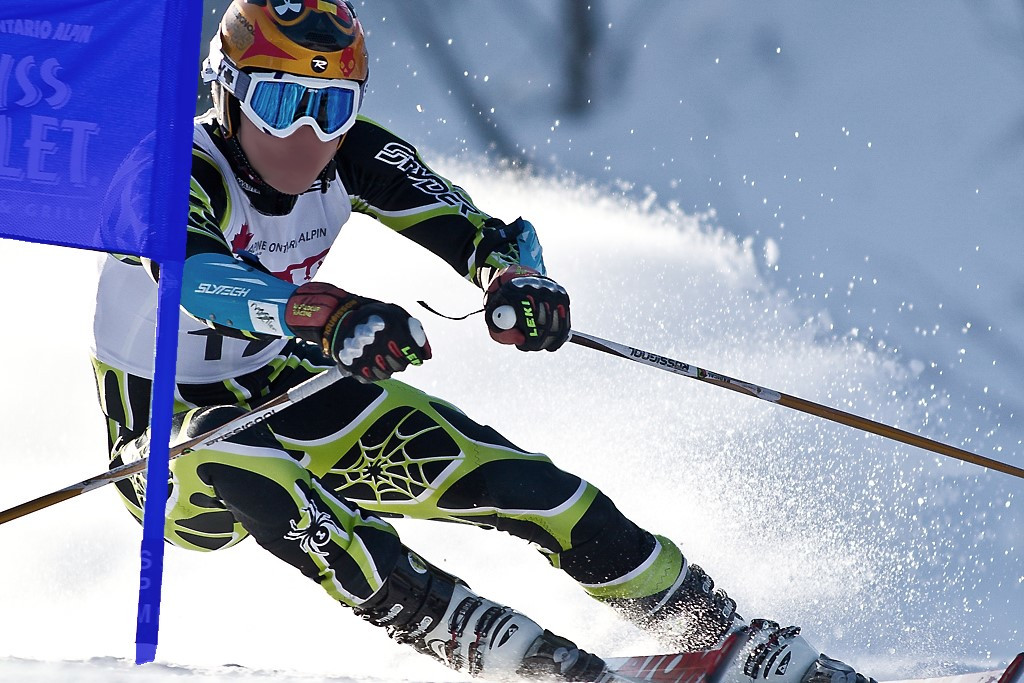} \\
         \includegraphics[width=\plotwidth]{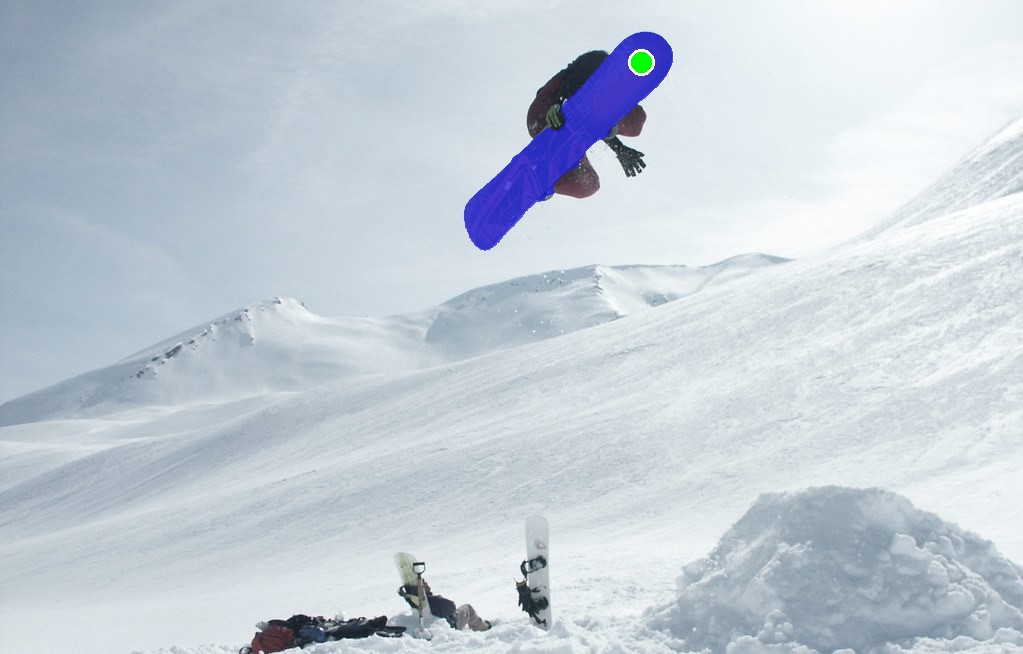} & 
         \includegraphics[width=\plotwidth]{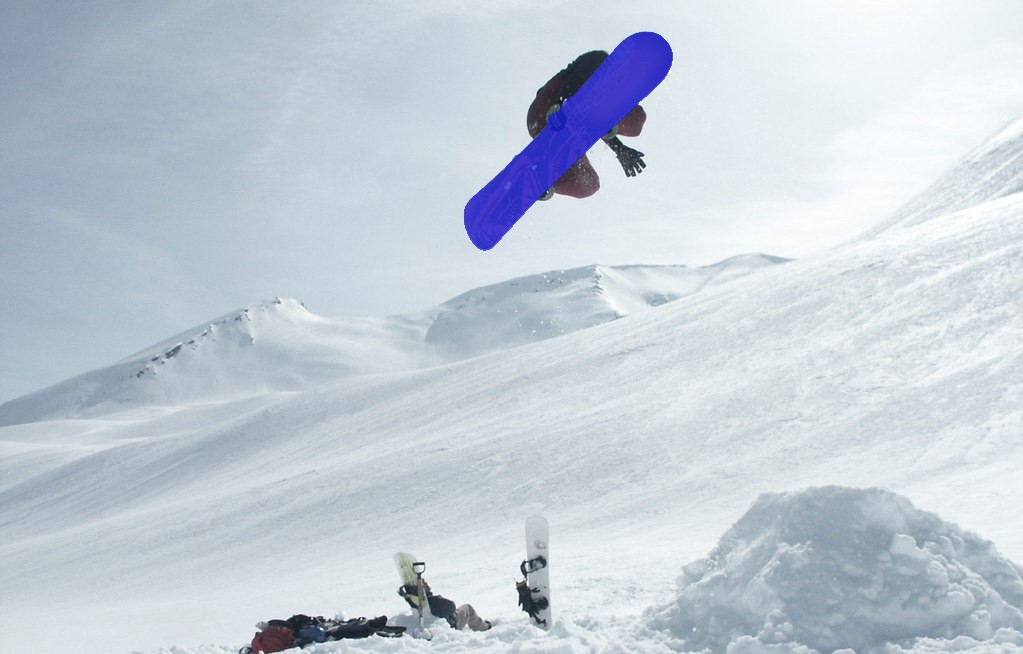} \\
         \includegraphics[width=\plotwidth]{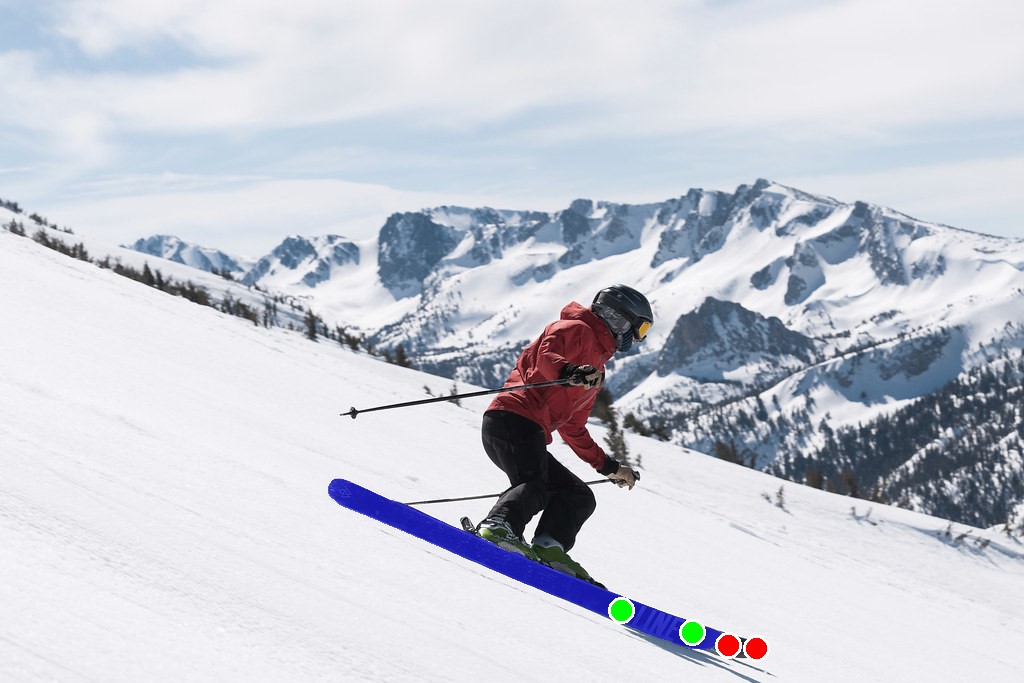} & 
         \includegraphics[width=\plotwidth]{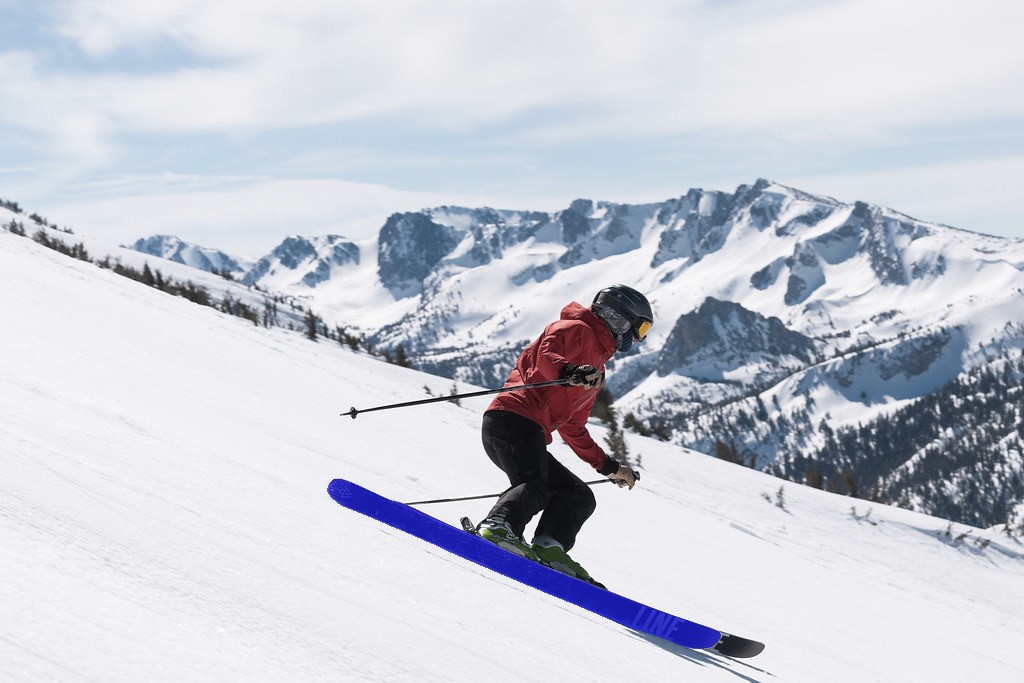} \\
         \includegraphics[width=\plotwidth]{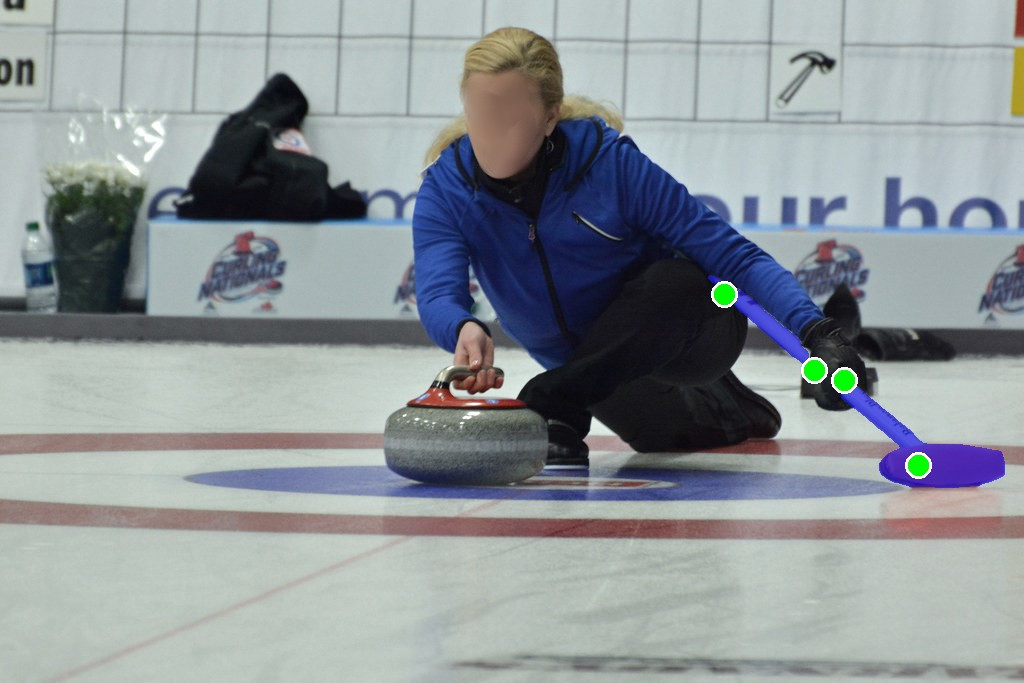} & 
         \includegraphics[width=\plotwidth]{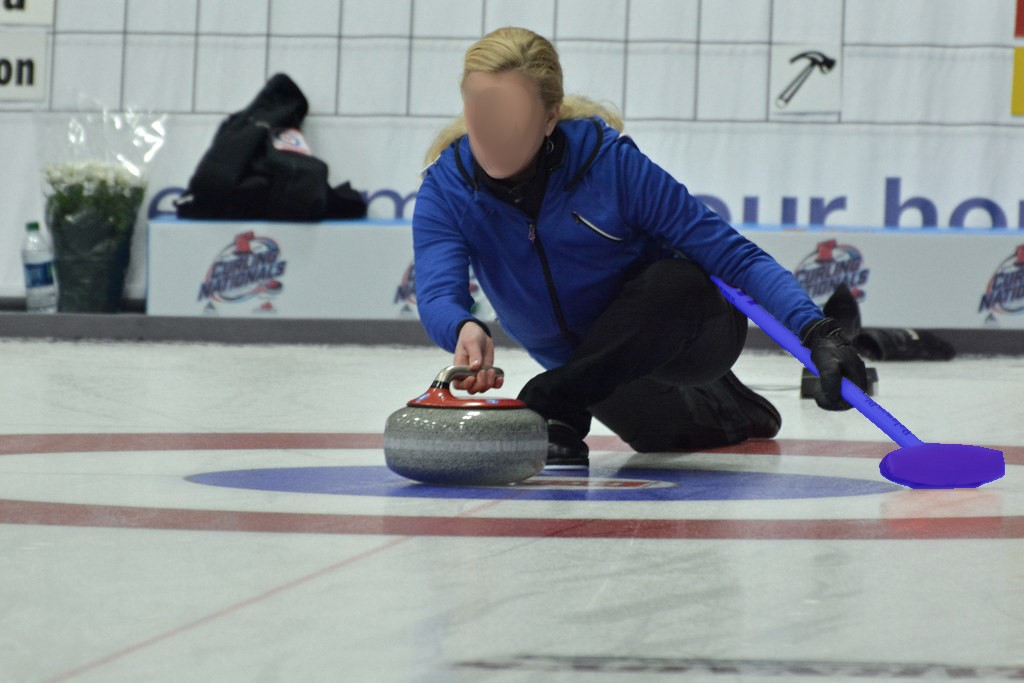} \\
         \includegraphics[width=\plotwidth]{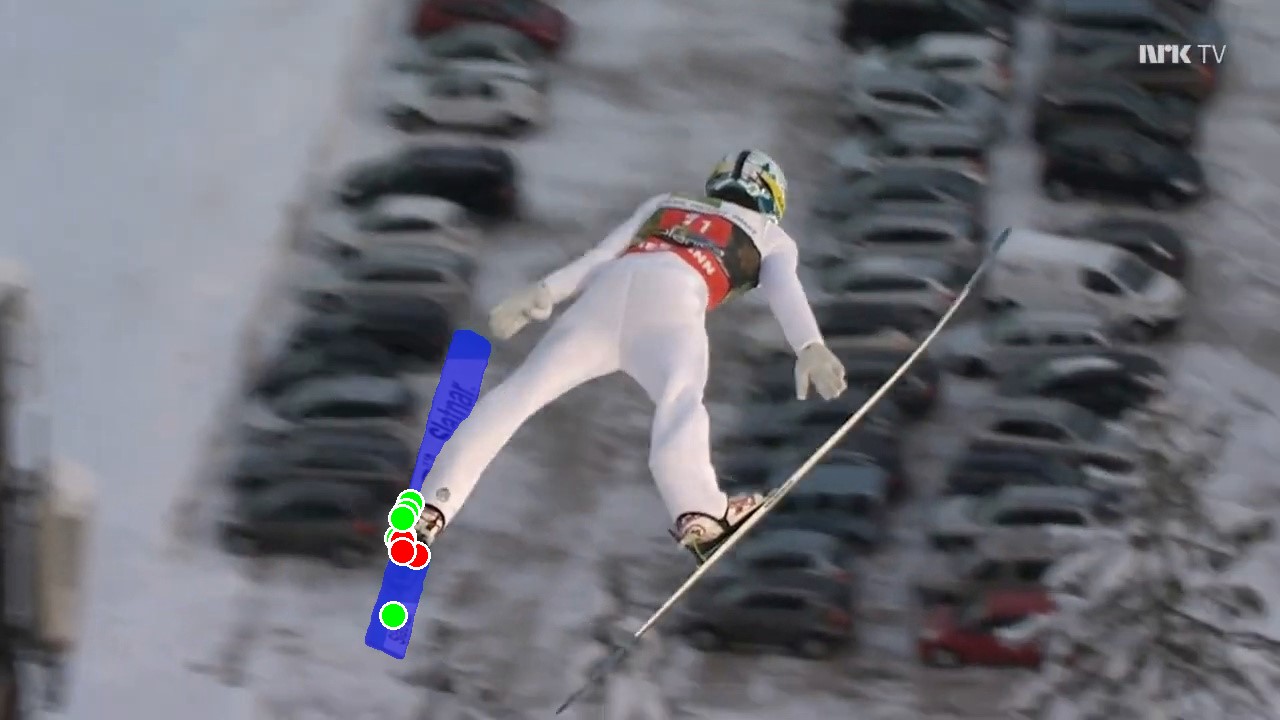} & 
         \includegraphics[width=\plotwidth]{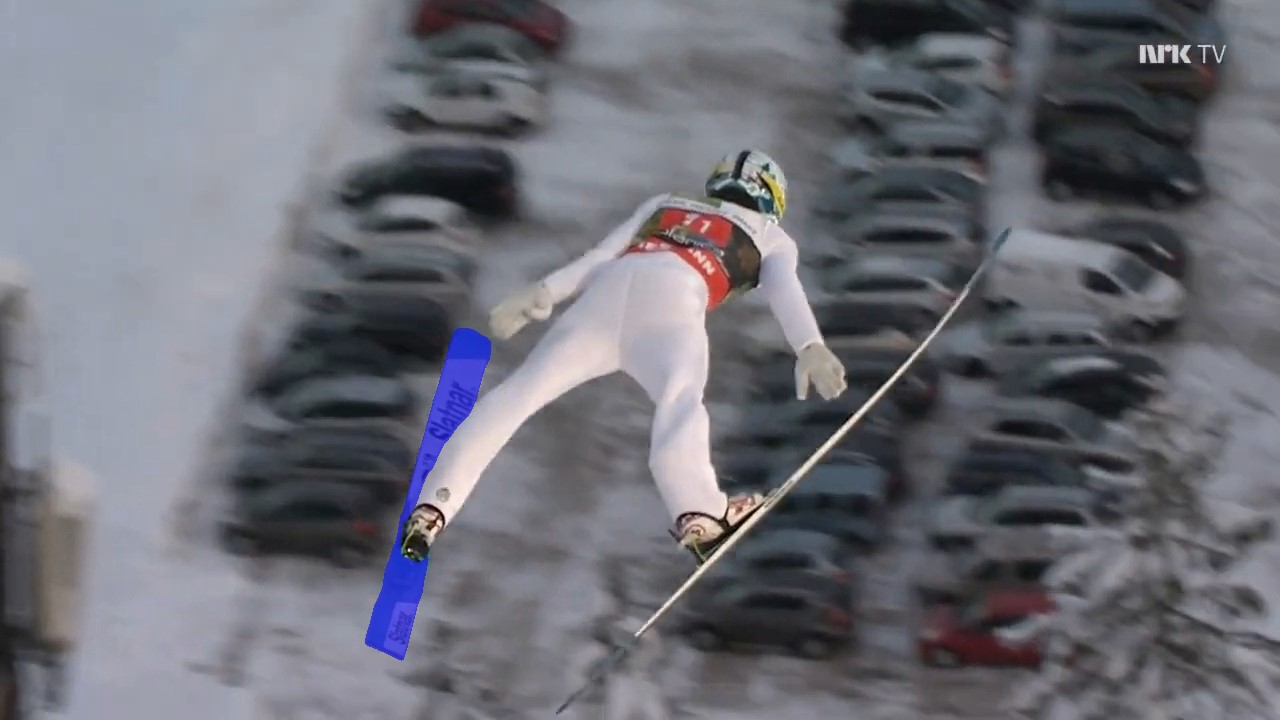} \\
         \includegraphics[width=\plotwidth]{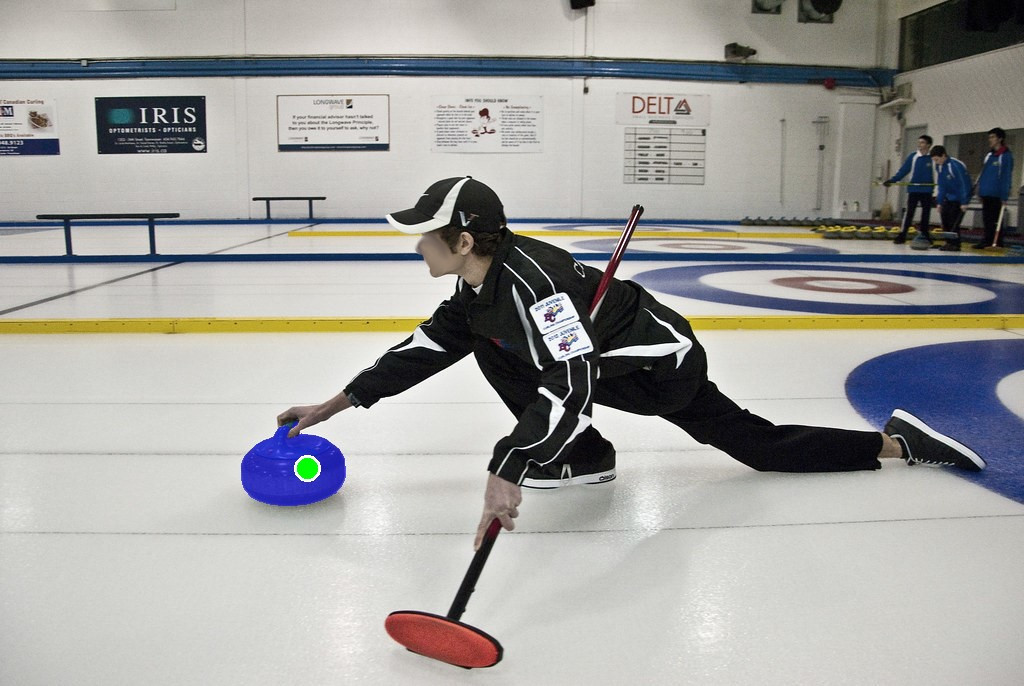} & 
         \includegraphics[width=\plotwidth]{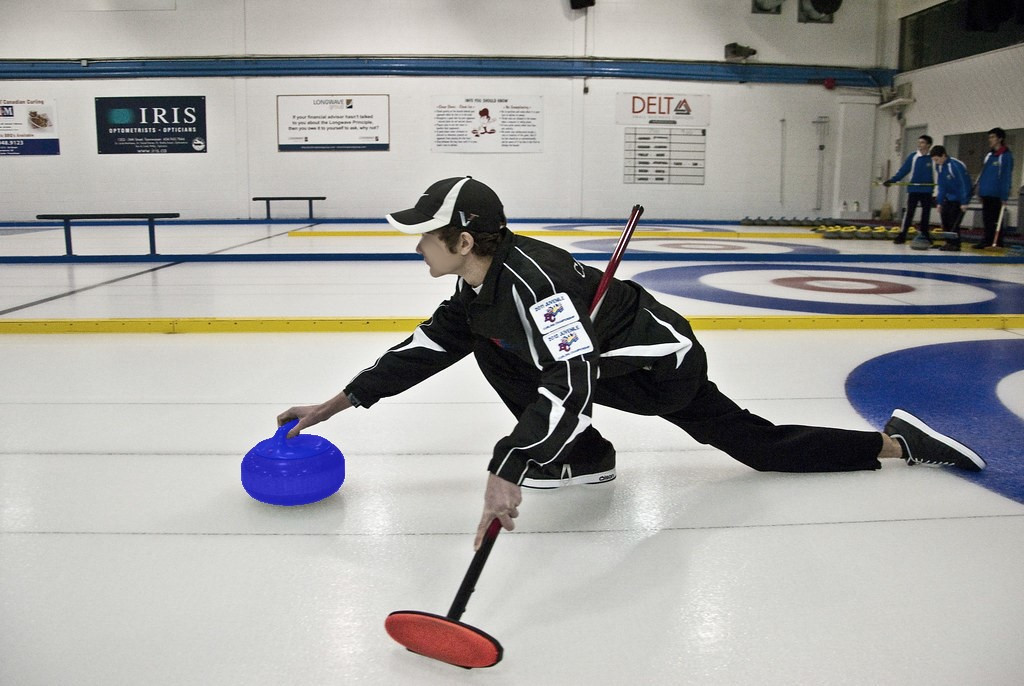} \\
         \includegraphics[width=\plotwidth]{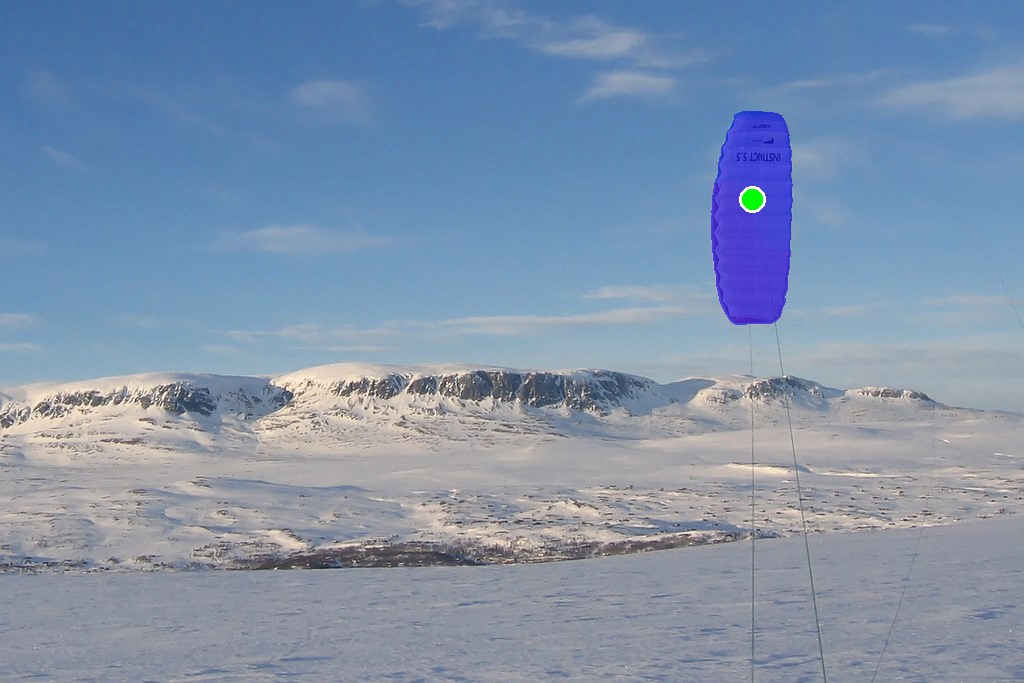} & 
         \includegraphics[width=\plotwidth]{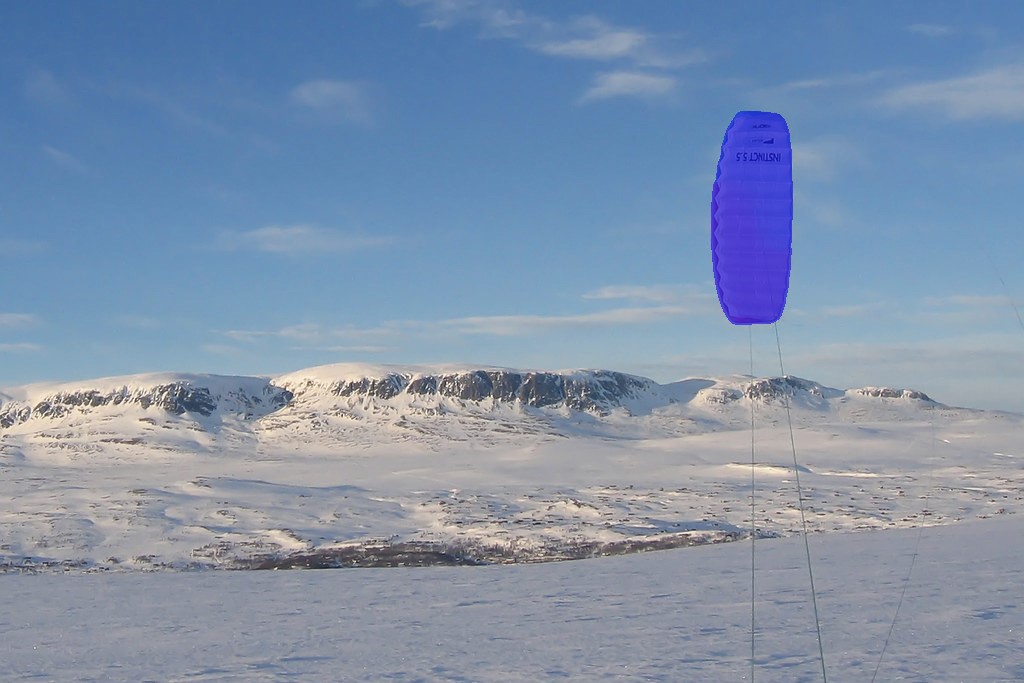} \\
    \end{tabular}
    }
    \caption{Qualitative examples on WSESeg. Foreground clicks are \emph{green}, background clicks are \emph{red} and the masks are \emph{blue}. \label{fig:qualitative}}
\end{figure}

\subsection{Experimental Setting} 
\label{sec:setting}
\paragraph{Metric. }
As mentioned in section \ref{subsec:problem}, the main indicator of the quality of an interactive segmentation system is number of clicks necessary to create a high quality mask. We will test our system on datasets with pre-existing ground truth masks $\mathbf{m}_\textbf{GT}$. First, we fix an IoU threshold $\theta_\textbf{IoU}$. A predicted mask $\mathbf{m}_\tau$ will be considered of sufficient quality if its IoU with $\mathbf{m}_\textbf{GT}$ reaches this threshold as in 
\begin{equation}
    \text{IoU}(\mathbf{m}_\tau, \mathbf{m}_\text{GT}) = \frac{|\mathbf{m}_\tau \cap \mathbf{m}_\text{GT}|}{|\mathbf{m}_\tau \cup \mathbf{m}_\text{GT}|} \ge \theta_\textbf{IoU}. 
\end{equation} 
The \underline{n}umber \underline{o}f \underline{c}licks (NoC) necessary to create such a sufficient mask is the most important metric in interactive segmentation. In some cases the system is effectively incapable to segment the mask, which might incur a potentially unbounded number of simulated interactions. To account for this problem, the maximum number of interactions is capped to 20. The resulting metric is called $\text{NoC}_{20}@ \theta_\textbf{IoU}$. 
Most notably, a lower NoC metric value indicates a better performance.
We follow \cite{ritm2022} for the simulation of user clicks. A detailed description of the clicks sampling algorithm can be found in the supplementary material. 

\paragraph{Implementation Details. } 
In all of our experiments we use a Adam optimizer \cite{kingma2017adammethodstochasticoptimization} with a learning rate of $\mu = 5 \cdot 10^{-5}$ and $\beta_1 = 0.9, \beta_2 = 0.999$. For all experiments we use focal loss \cite{ross2017focal}. 
We train on a combination of COCO and LVIS \cite{lin2014microsoft, gupta2019lvis} for 55 epochs. It should be noted that one epoch is defined as 30,000 images. 
We downscale our learning rate by $\frac{1}{10}$ after epoch 50 and 55. 
We follow the practice of \cite{Liu_2023_ICCV} and initially sample up to 24 random points. Iterative training with simulated points is carried out for three additional iterations. 
Our backbone is a ViT-B \cite{DBLP:conf/iclr/DosovitskiyB0WZ21} that has been pretrained with DINOv2 \cite{oquab2023dinov2}. 
During training we use random crops of size $448 \times 448$. We also rescale our images to this during testing unless said otherwise. For the WSESeg, SHSeg and HQSeg-44k datasets we use a resolution of $896 \times 896$ in order be able compare our system to SAM and HQ-SAM, which use a resolution of $1024 \times 1024$. For the DAVIS dataset we use a resolution of $672 \times 672$. All our resolutions stem from the requirement of being multiple of the patch size $14 \times 14$. 

\paragraph{Datasets. }
We follow common practice and use the COCO+LVIS dataset for training. This dataset constitutes a combination of the COCO dataset \cite{lin2014microsoft} along with its annotations, and the LVIS dataset \cite{gupta2019lvis} that has been introduced in \cite{ritm2022}. 
Our main ablation study is carried out on the WSESeg dataset for the segmentation of winter sports equipment. This dataset contains 7452 masks for ten different classes of objects. We also carry out some evaluations on datasets which are commonly used in interactive segmentation: GrabCut \cite{grabcut2004}, Berkeley \cite{MartinFTM01}, DAVIS \cite{Perazzi_2016_CVPR} and SBD \cite{hariharan2011semantic}. 
In addition to this, we make use of the HQSeg-44k dataset \cite{gupta2019lvis}, which effectively is a mix of various high-quality human-annotated datasets. Especially DAVIS and HQSeg-44k allow us to test the efficacy of the architectural modifications on data which contains fine-grained details. 

\paragraph{The SHSeg dataset. }
We present a novel dataset containing masks for skiing humans, which we call \emph{SHSeg} for \emph{\underline{S}kiing \underline{H}uman \underline{Seg}mentation}. Since WSESeg mostly focuses on the segmentation of the equipment in winter sports contexts, this data allows to test our models capacity to segment the athletes themselves. 
The images themselves originate from the SkiTB dataset \cite{dunnhofer2024tracking, SkiTBcviu}, and are already annotated with one bounding box of the skier per image. In order to select the images we use, we first filter out all images whose corresponding bounding box has a height or width of less than 150 pixels. Out of the remaining images, we randomly sample 500 images. We only annotate images that show a clearly distinguishable skier, which leaves us with 496 images. Since some images contain more than one athlete, we end up with 534 masks in total. We annotated the dataset with the system provided in \cite{Liu_2023_ICCV}, since using our own model would have introduced a bias. It should be noted that, while the annotations are created using interactive segmentation systems, all annotations are inspected and corrective clicks are provided by a human until a very high degree of quality is reached.

\subsection{Results}

\begin{table}[]
    \centering
    \resizebox{0.80\linewidth}{!}{
    
    \begin{tabular}{|l|c|c|c|c|c|}
        \hline 
        \multirow{2}{*}{\textbf{Configuration}} & GrabCut & Berkeley \\
        & NoC@90 & NoC@90 \\ 
        \hline
        Baseline & 1.74 & 3.86 \\
        + Frozen Backbone & 1.72 & 3.73 \\ 
        + Intermediate Features & \textbf{1.40} & 2.84 \\ 
        + Skip Connections & 1.44 & \textbf{2.45} \\
        \hline 
        DAVIS & SBD & HQSeg-44k\\
        NoC@90 & NoC@90 & NoC@90 \\
        \hline 
        6.75 & 8.53 & 7.91 \\
        6.40 & 8.51 & 7.98 \\ 
        5.05 & 6.65 & 6.40 \\ 
        \textbf{4.94} & \textbf{6.18} & \textbf{6.00} \\
        \hline 
    \end{tabular}
    }
    \caption{Ablation study on general consumer images. A lower NoC metric indicates better performance.}
    \label{tab:general_ablation}
\end{table}

\begin{table*}[t]
    \centering
    \resizebox{0.8\linewidth}{!}{
    
    \begin{tabular}{|l|l|c|c|c|c|c|c|}
    \hline
         \multirow{2}{*}{\textbf{Method}} &\multirow{2}{*}{\textbf{Fusion type}} & \multicolumn{3}{c|}{DAVIS} & \multicolumn{3}{c|}{HQSeg-44K} \\
         & & NoC@90 & NoC@95 & $\ge20@95$ & NoC@90 & NoC@95 & $\ge20@95$ \\
         \hline 
         RITM \cite{ritm2022} & \textit{early} & 5.34 & 11.45 & 139 & 10.01 & 14.85 & 910 \\
         FocalClick \cite{chen2022focalclick} & \textit{early} & \textbf{4.90} & 10.40 & 123 & 7.03 & 10.74 & 649 \\ 
         SimpleClick \cite{Liu_2023_ICCV} & \textit{early} & 5.06 & 10.37 & \textbf{107} & 7.47 & 12.39 & 797 \\
         \hline 
         SAM \cite{kirillov2023segment} & \textit{late} & 5.14 & 10.74 & 154 & 7.46 & 12.42 & 811 \\
         MobileSAM \cite{mobile_sam} & \textit{late} & 5.83 & 12.74 & 196 & 8.70 & 13.83 & 951 \\ 
         HQ-SAM  \cite{ke2024segment} & \textit{late} & 5.26 & \textbf{10.00} & 136 & 6.49 & 10.79 & 671 \\ 
         SegNext \cite{liu2024rethinking} & \textit{late} & 5.34 & 12.80 & 163 & 7.18 & 11.52 & 700 \\ 
         InterFormer \cite{huang2023interformer} & \textit{late} & 5.45 & 11.88 & 150 & 7.17 & 10.77 & 658 \\ 
         \hline
         SkipClick (ours) & \textit{late} & 4.94 & 11.92 & 158 & \textbf{6.00} & \textbf{9.89} & \textbf{608} \\
         \hline
    \end{tabular}
    }
    \caption{Segmentation of images with fine granularity in their segmentation masks. Our SkipClick method achieves state-of-the-art results on HQSeg-44k, which corroborates the general applicability of our method. $\ge20@95$ denotes the number of masks for which the test run needed at least 20 clicks to attain a IoU of at least 95 with the pre-existing ground truth. For all metrics, lower is better. Results for other methods are from \cite{liu2024rethinking}. }
    \label{tab:high_detail}
\end{table*}

The design aspects of our architecture have already been validated for winter sports in \cref{tab:wseseg_ablation} in \cref{sec:method_construction}. 
In \cref{tab:wseseg_comparison,tab:shseg_comparison} we compare ourselves with SAM \cite{kirillov2023segment}, HQ-SAM \cite{ke2024segment} and the method by \cite{schoen2024wseseg} which has been applied to the two architectures.  
The SAM (Segment Anything Model) architecture introduced the strategy of late fusion in interactive segmentation and has been trained on the largest interactive segmentation dataset to date: The SA-1B dataset, which has been published alongside the model and contains 1.1B masks on 11 million images. 
Despite such an extensive amount of training data, we observe an inferior performance when compared to our SkipClick model, which has been trained on the vastly smaller COCO+LVIS dataset (99k images with 1.5 million masks). SAM achieves a NoC@85 of 8.83 in comparison to our SkipClick model with a NoC@85 of 6.49. The slightly more challenging NoC@90 metric supports this observation with 11.86 for SAM and 9.16 for SkipClick. 
We mainly attribute the performance improvement to the fact that our model makes extensive use of low-level features, while SAM only uses the tensor produced by the last layer of its image encoder. 
The latter hypothesis is especially supported by looking at our baseline in \cref{tab:wseseg_ablation}, which does not use intermediate features and only achieves an average NoC@85 of 9.46. The baseline is thus worse than SAM, while the usage of intermediate features results in a model that is better than SAM. 
Our model outperforms HQ-SAM on all classes. 
Even when accounting for the improvements incurred by the method in \cite{schoen2024wseseg}, on average, our model still outperforms both SAM and HQ-SAM. 
\Cref{fig:radar_plot_comparison} displays our comparison on WSESeg as a radar plot. Examples for qualitative results can be found in \cref{fig:qualitative}. 
The results on our novel SHSeg dataset corroborate our comparison. Our SkipClick model reduces the NoC@85 metric by 62.8 \% in comparsion to the second best method (SAM + Schön et al.) and the NoC@90 metric by 66.2 \% in comparison to SAM. 
Qualitative examples on the SHSeg dataset can be found in the supplementary material. 
We also measured the speed of the methods in \cref{tab:wseseg_comparison} in terms of the time the model needs to respond after each click. The slowest methods are the modifications of SAM and HQ-SAM provided by Schön et al. \cite{schoen2024wseseg}, which have a response time of 41.38 ms and 56.13 ms, respectively. This is caused by the necessity of an additional backpropagation in their framework. The standard SAM and HQ-SAM need 15.01 ms and 18.83 ms to respond. Out of the compared methods, ours is the fastest with a response time of 6.61 ms. All these durations have been measured using a V100 GPU.

In \cref{tab:general_ablation}, we also evaluate our main ablation study on datasets that comport with general consumer images. The results on these datasets generally support the results obtained during the method construction (seen in \cref{tab:wseseg_ablation}). In almost all cases our complete methods delivers the strongest results except for one: The skip connections slightly worsen the performance on GrabCut. We assume this to be the case due to a slight overfitting to particularities of COCO+LVIS in comparison to the GrabCut dataset.

We also compare our method to existing methods on the DAVIS and HQSeg-44k datasets in \cref{tab:high_detail}. Both datasets are characterized by a high level of detail in their masks. With respect to the NoC@90 metric on DAVIS, our model performs best when compared to the other late fusion models with 4.94 clicks on average. If we also compare to the early fusion models, we are only second best with FocalClick having a NoC@90 of 4.90. On HQSeg-44k our model achieves state-of-the-art results for all explored metrics. The results in \cref{tab:general_ablation,tab:high_detail} demonstrate that our architectural design has not been overfit to the winter sports domain, but instead performs also well on other domains.

\section{Conclusion}
In our paper we presented a new architecture for interactive segmentation on the winter sports domain. We first describe a baseline where we pay particular attention to equip the model with the capacity to deliver real-time responses after each clicks, by employing late fusion. Afterwards, we add certain architectural features for which we are able to show that they improve the performance when tasked with segmenting winter sports equipment. 
We are able to show that our model outperforms SAM and HQ-SAM on the WSESeg dataset. We manage to confirm this result on a newly published dataset for the segmentation of skiers, called SHSeg. Additionally, when carrying out our ablation study on general consumer image datasets, we are able to show that our model's architecture is not overfit to the domain of winter sports equipment. When comparing our method to existing methods, we achieve competitive results on the DAVIS dataset and state-of-the-art results on the HQSeg-44k dataset.

{\small
\bibliographystyle{ieee_fullname}
\bibliography{egbib}
}

\appendix

\section{Simulating User Clicks}
In this section we discuss how we simulate the clicks. More precisely, we want to answer the question: Given a predicted mask $\mathbf{m}_\tau$ and a ground truth mask $\mathbf{m}_\text{GT}$, where do we place the next click in order to help our network with improving the mask? We follow common practice and use the method described in \cite{ritm2022}. 
\begin{enumerate}
    \item For each click we simulate, we compare $\mathbf{m}_\tau$ with $\mathbf{m}_\text{GT}$ to obtain the mask of false positives $\mathbf{m}_\text{FP}$ and the mask of false negatives $\mathbf{m}_\text{FN}$. 
    \item Afterwards, we compute the euclidean distance transforms (see \cite{felzenszwalb2012distance}) of both masks, $\mathcal{D}_\text{FP}$ and $\mathcal{D}_\text{FN}$. 
    \item We will then look for the maxima of $\mathcal{D}_\text{FP}$ and $\mathcal{D}_\text{FN}$. The coordinates of the higher maximum will be the location of the simulated click. 
    \item Depending on whether this maximum is found in either $\mathcal{D}_\text{FP}$ or $\mathcal{D}_\text{FN}$, we will label it as a background (-) or foreground (+) click, respectively.
\end{enumerate}

It should be noted that this metric allows for improving the system at the cost of practical usability. If we were to simulate the clicks during training in the exact same way as we do during testing (taking the maximum of the two distance transforms), we would prepare our model to optimally perform under the metric. This can for example be seen in \cite{Lee_2024_CVPR, huang2023interformer}. As \cite{ritm2022} however mentions, this inhibits the practical usability of the model, since an actual human would choose other non-optimal click positions. We would see a kind of overfitting to the metric. To make sure the training of our model adheres to practical requirements, we follow common practice \cite{ritm2022, Liu_2023_ICCV, kirillov2023segment, chen2022focalclick, pseudoclick2022} and use additional random clicks during each training step.

\section{Changing the Number of Encoder Blocks after Adding the Prompts}
The prompt features and the image features in our architecture are fused by multiple transformer encoder blocks. In our standard model we chose four as the number of blocks. 
In \cref{tab:depth_comparison} we compare the performance of the model when altering the number of blocks (the column \emph{Depth}). We cannot observe a clear trend, as a continuous increase of the number of blocks does not necessarily cause an improvement. We even see that reducing the number of blocks to three gives a slightly better performance for a NoC@85 of 6.94 to 6.31, although the best performance depends on the metric, with 6.31 for the NoC@85 and 9.023 for the NoC@90.

\begin{table}[t]
    \centering
    \begin{tabular}{|c|c|c|}
         \hline
         \multirow{2}{*}{Depth} & \multicolumn{2}{c|}{WSESeg Average} \\
         & NoC@85 & NoC@90 \\
         \hline 
         2 & 6.962 & 9.587 \\
         3 & \textbf{6.311} & 9.091 \\
         4 & 6.944 & 9.163 \\
         5 & 6.689 & 9.310 \\ 
         6 & 6.524 & \textbf{9.023} \\
         \hline
    \end{tabular}
    \caption{A comparison of the change in performance for different numbers of ViT blocks. The \emph{depth} does not refer to the backbone, but the additional blocks after mixing the image and prompt features. The NoC is the average over all classes. }
    \label{tab:depth_comparison}
\end{table}

\section{Qualitative Examples from SHSeg}
\begin{figure*}
    \centering
    {\def\plotwidth{0.4\linewidth}
    \setlength{\tabcolsep}{1.5pt}
    \begin{tabular}{cc} 
         \textbf{Prediction} & \textbf{Ground Truth} \\
         \includegraphics[width=\plotwidth]{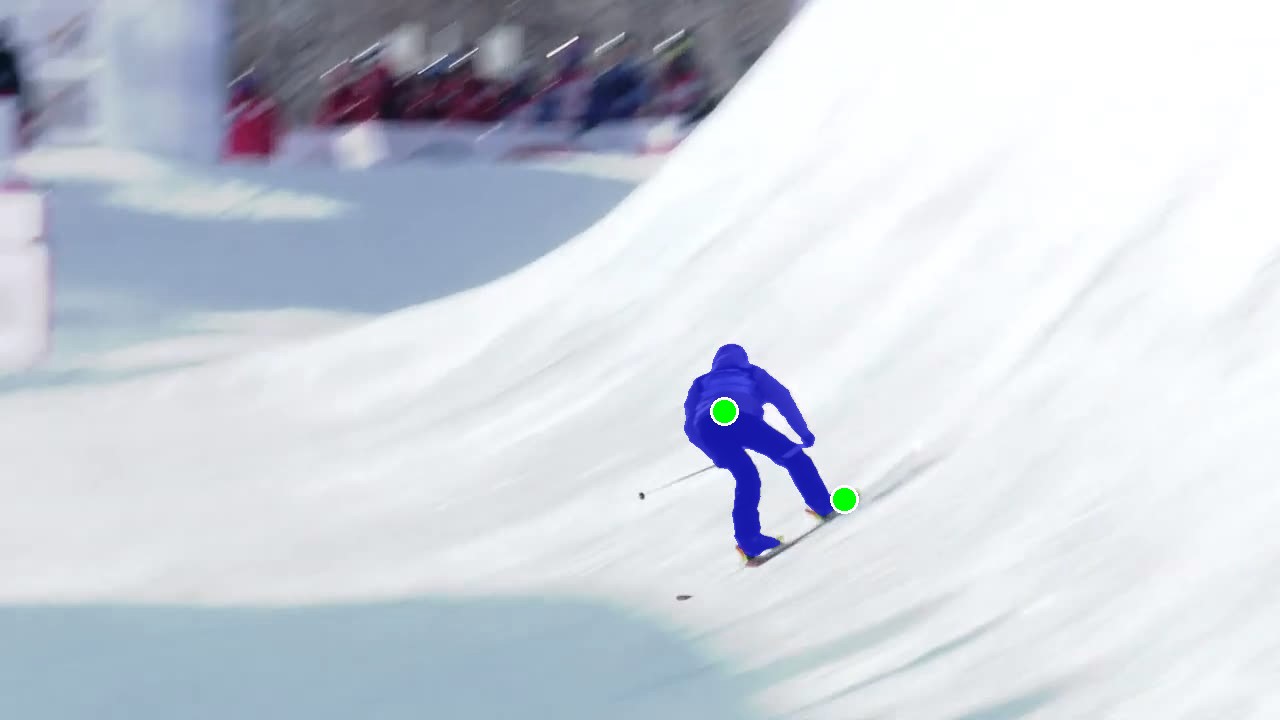} & 
         \includegraphics[width=\plotwidth]{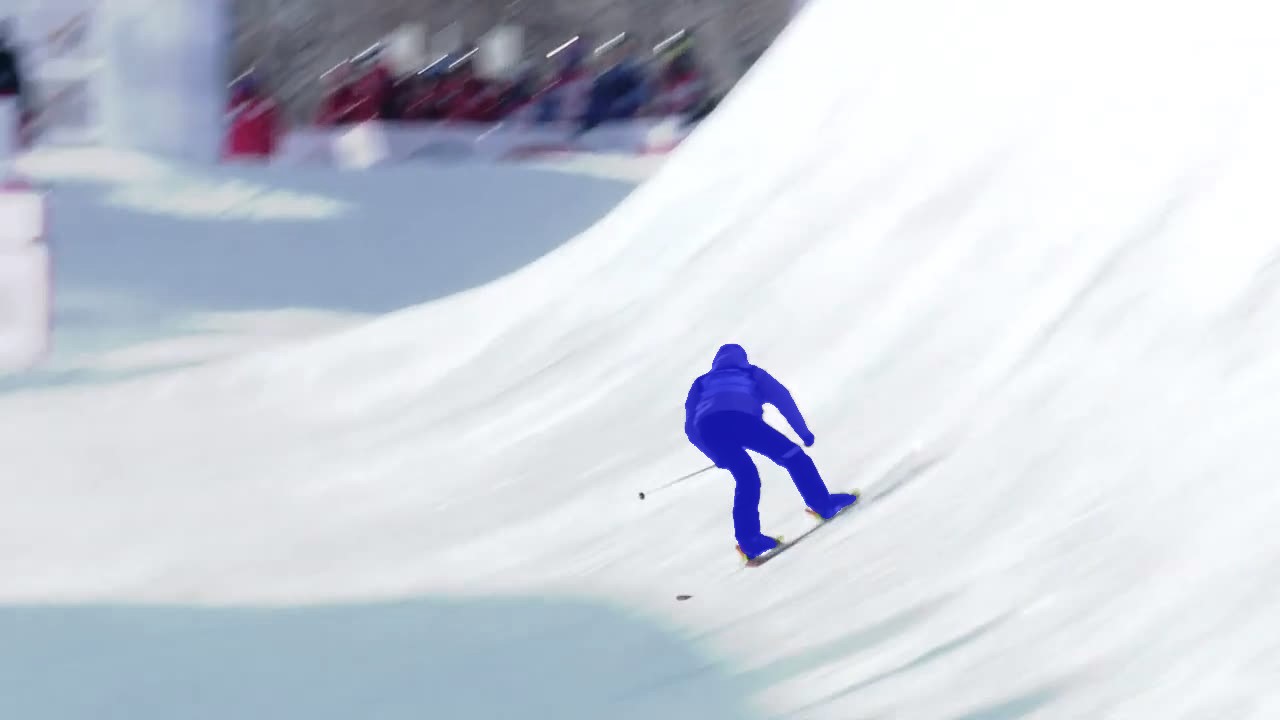} \\
         \includegraphics[width=\plotwidth]{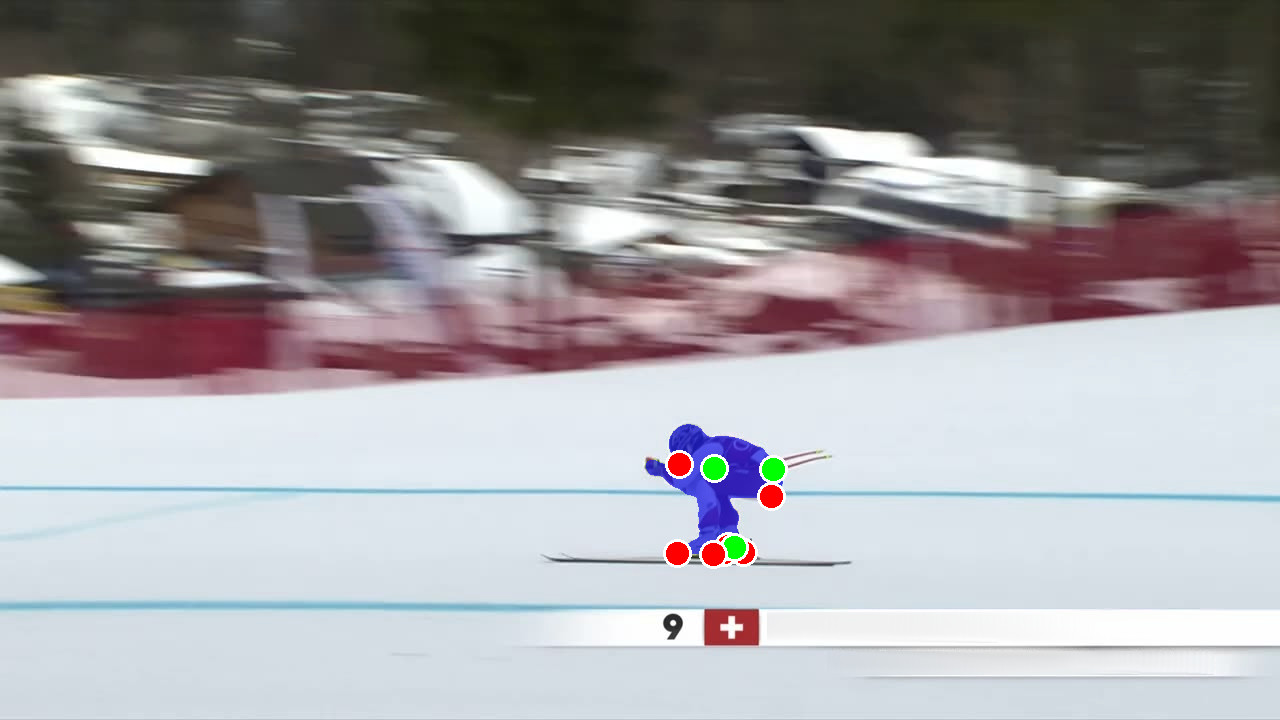} & 
         \includegraphics[width=\plotwidth]{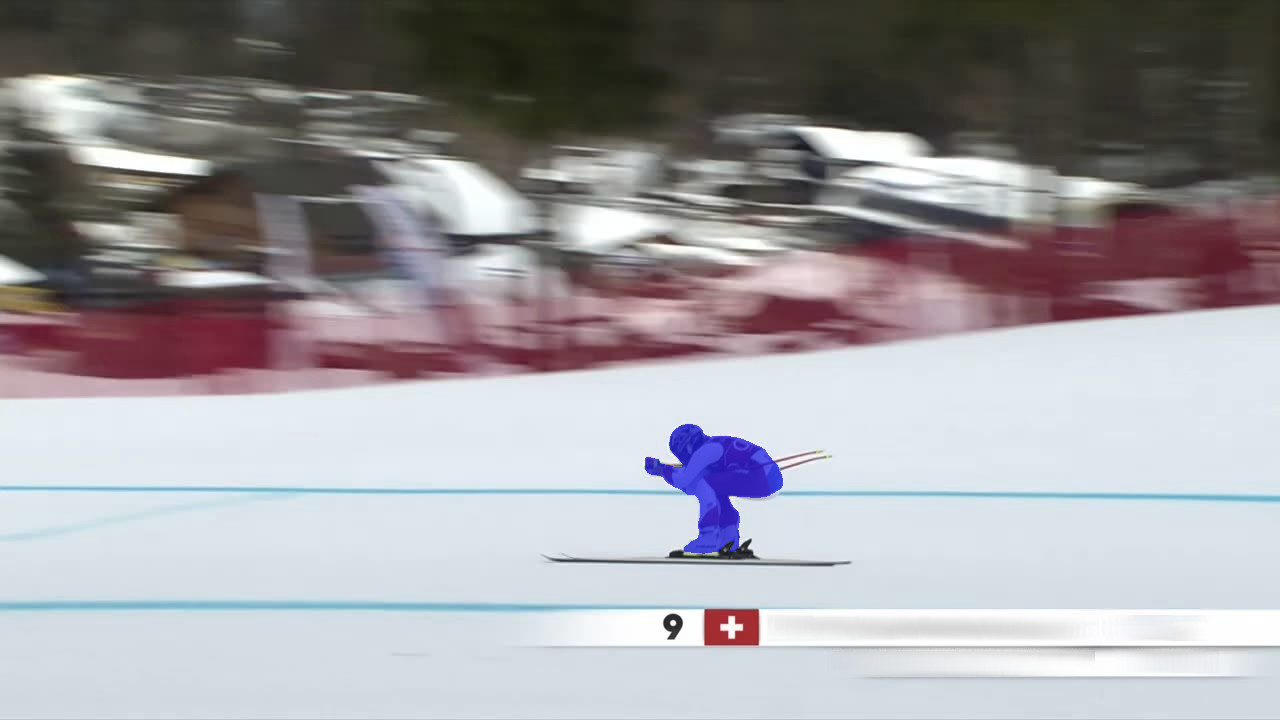} \\
         \includegraphics[width=\plotwidth]{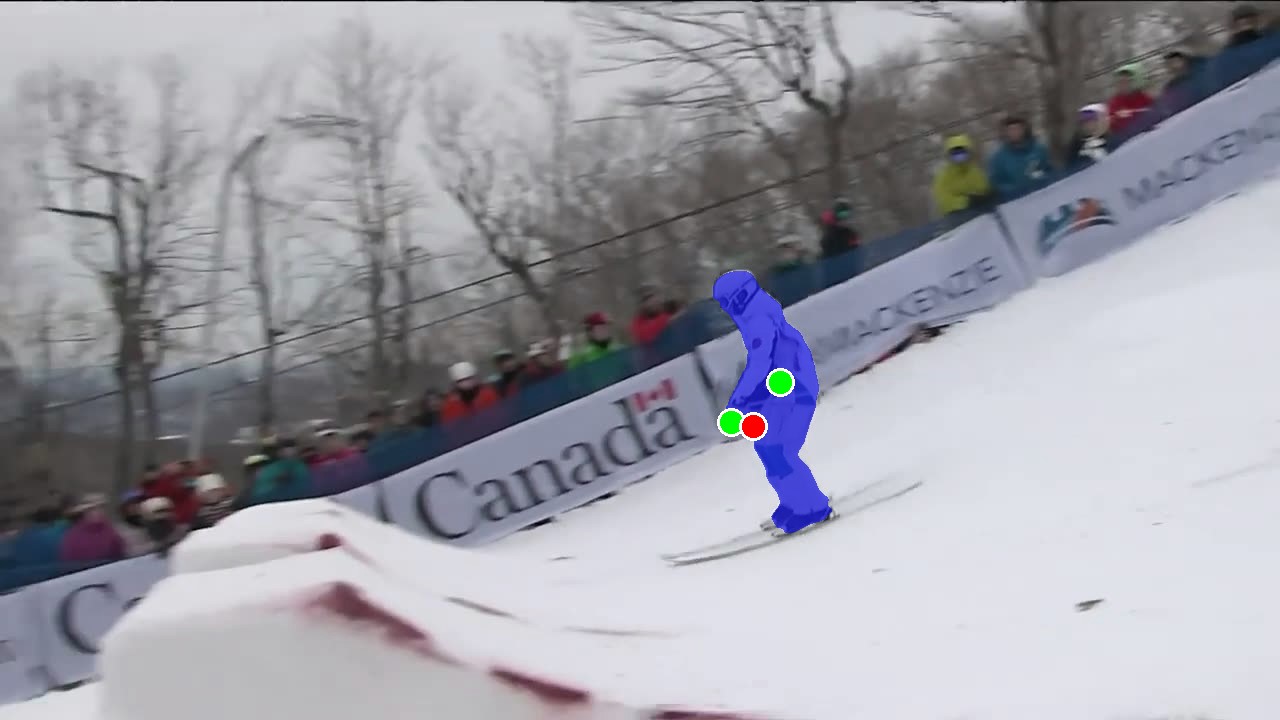} & 
         \includegraphics[width=\plotwidth]{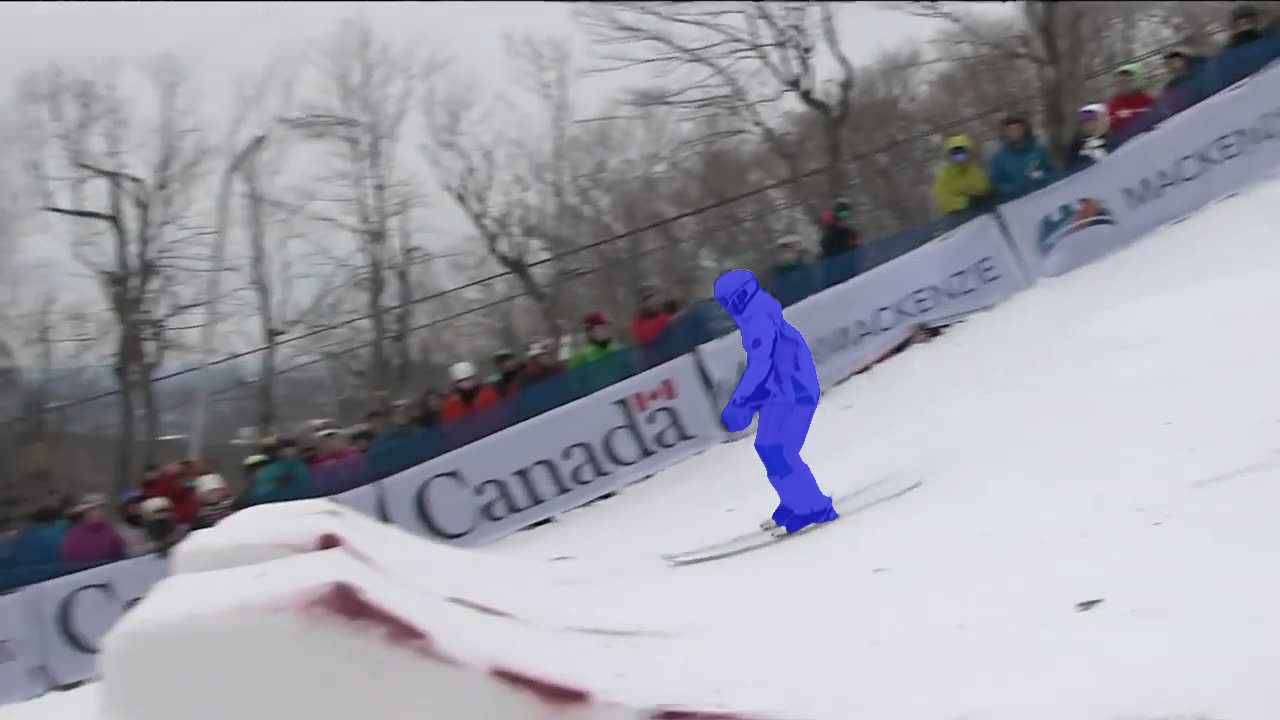} \\
         \includegraphics[width=\plotwidth]{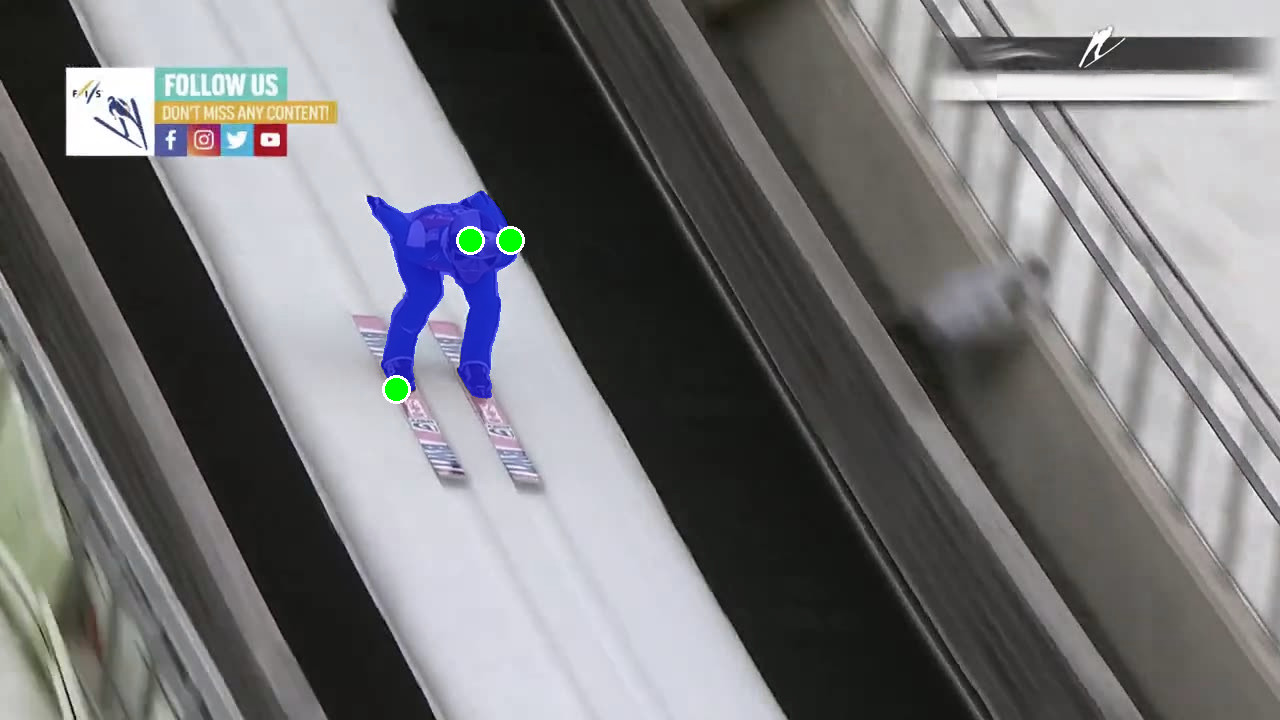} & 
         \includegraphics[width=\plotwidth]{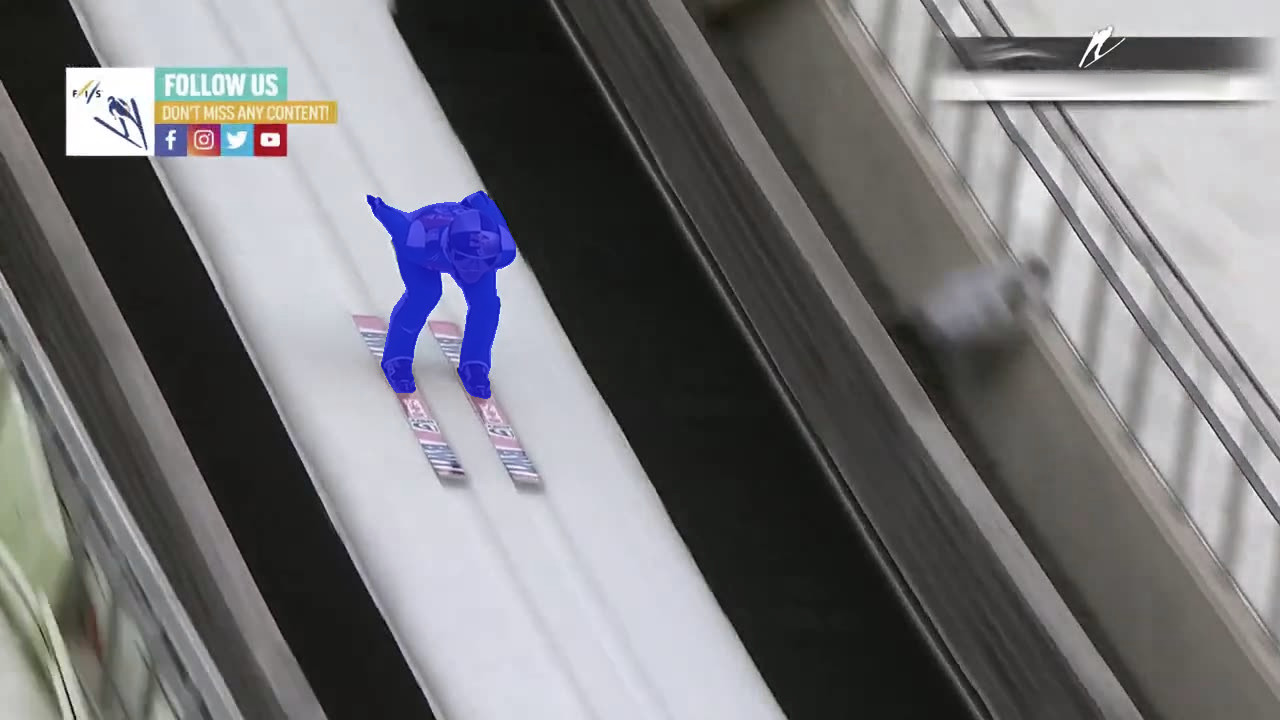} \\
         \includegraphics[width=\plotwidth]{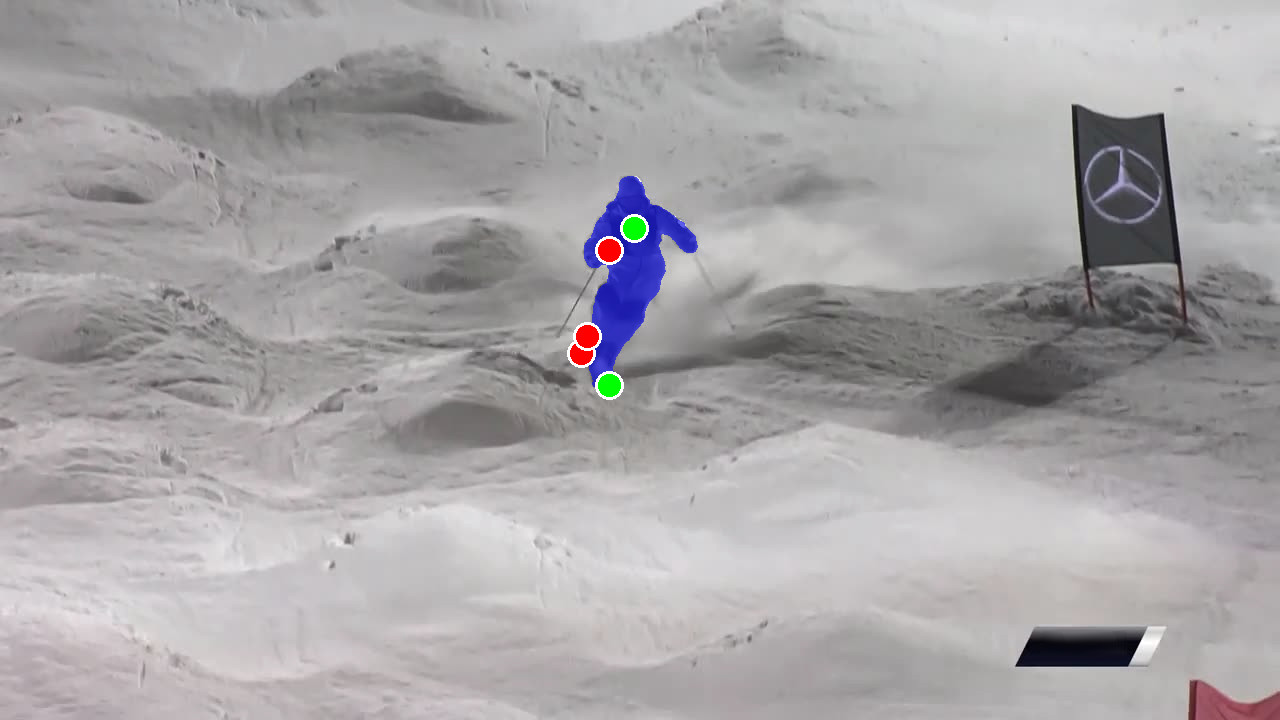} & 
         \includegraphics[width=\plotwidth]{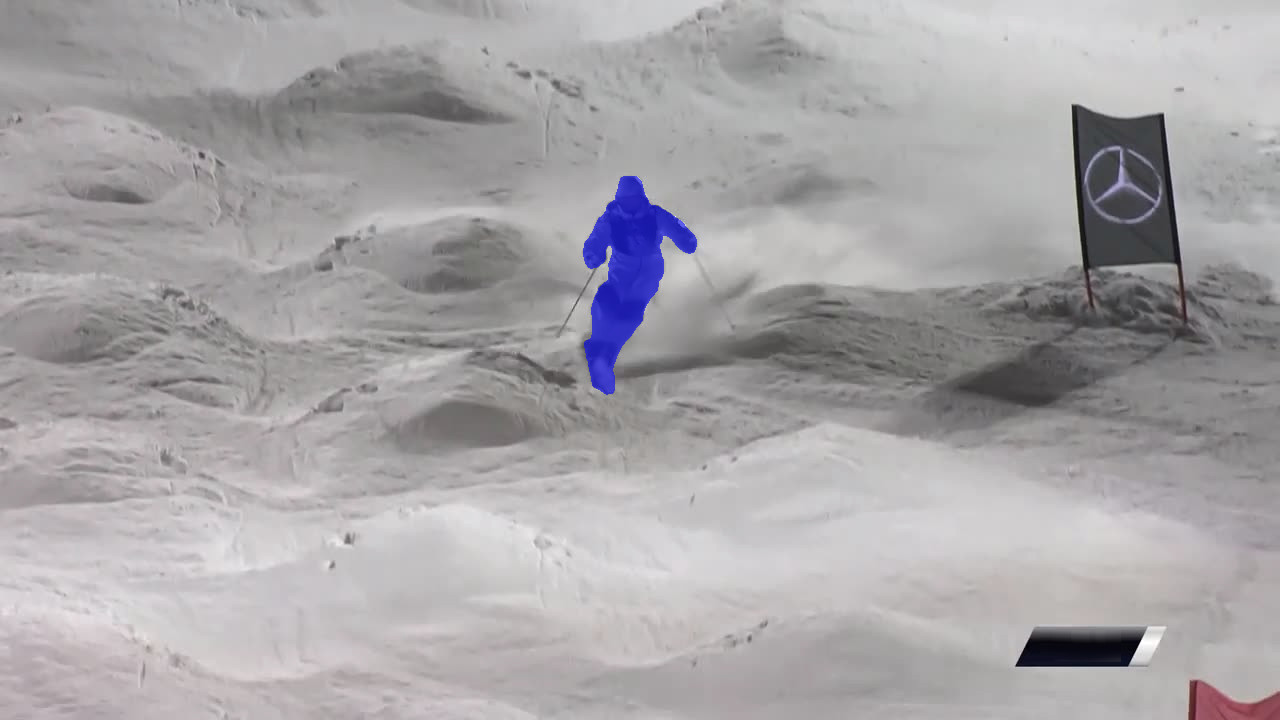} \\
    \end{tabular}
    }
    \caption{Examples for the masks occurring during the interaction. The \emph{left column} displays the predicted mask along with the clicks. Foreground clicks are \emph{green}, background clicks are \emph{red} and the masks are \emph{blue}. The \emph{right column} displays the corresponding ground truth. \label{fig:qualitative_shseg}}
\end{figure*}

In \Cref{fig:qualitative_shseg}, we can see qualitative examples from our newly proposed SHSeg (\underline{S}kiing \underline{H}uman \underline{Seg}mentation) dataset. Our dataset provides 534 masks for skiers on 496 images. The images have been randomly sampled from the SkiTB dataset \cite{dunnhofer2024tracking,SkiTBcviu}. A link to the data can be found in our main paper (publication of data upon acceptance). 

\end{document}